%% file: paper.tex
\documentclass[]{sjtu_sai}
\usepackage[toc,page,header]{appendix}
\usepackage{minitoc}
\usepackage{algorithm}
\usepackage{algpseudocode}
\usepackage{booktabs}
\usepackage{multirow}
\usepackage{siunitx}
\usepackage{makecell}
\usepackage{mathtools}
\usepackage{tabularx,array}
\usepackage[table]{xcolor}
\usepackage{graphicx}
\usepackage{wrapfig}
\usepackage{xspace}
\usepackage{amsmath}
\usepackage{amsfonts}

\usepackage{hyperref}
\usepackage{url}

\usepackage{booktabs}          
\usepackage{array}             
\usepackage[table]{xcolor}     
\usepackage{booktabs}
\usepackage{tabularx}
\usepackage{caption}
\usepackage{array}
\usepackage{multirow}
\usepackage[table]{xcolor}
\usepackage{graphicx}

\usepackage{wrapfig}
\usepackage{eucal}
\definecolor{grpbg}{HTML}{E3F0FA}   
\definecolor{oursbg}{HTML}{E6E6E6}  

\AtBeginEnvironment{tabular}{\renewcommand{\textbf}[1]{{\bfseries #1}}}
\AtBeginEnvironment{tabular*}{\renewcommand{\textbf}[1]{{\bfseries #1}}}
\AtBeginEnvironment{tabularx}{\renewcommand{\textbf}[1]{{\bfseries #1}}}

\title{SparseWAM: Driving World-Action Models are Worth 16 Register Tokens}

\titlelogo[trim=110bp 149bp 93bp 89bp,clip]{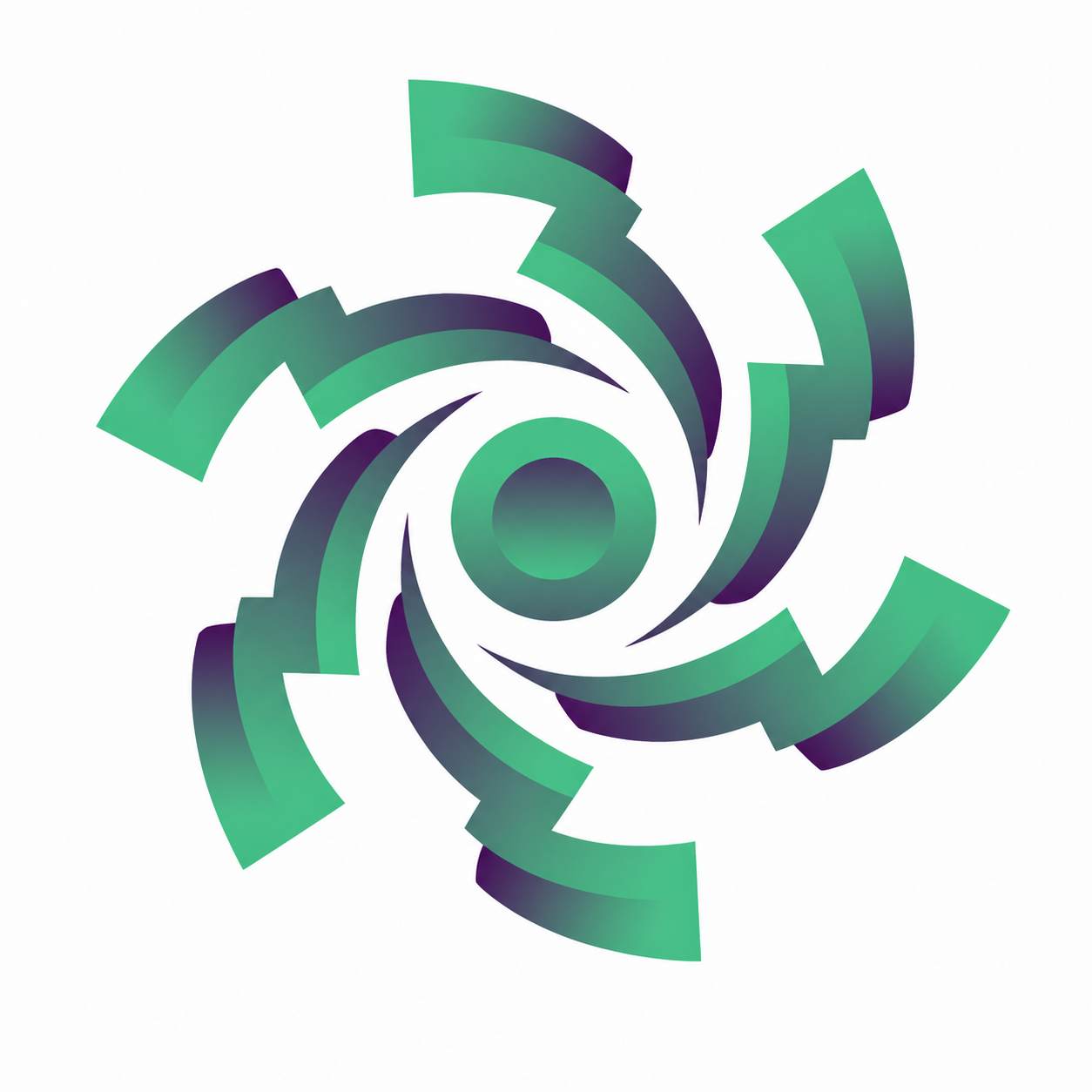}
\titlelogo[trim=6bp 4bp 4bp 6bp,clip]{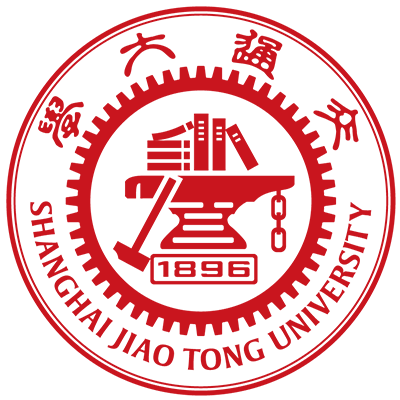}
\titlelogo[trim=2bp 2bp 2bp 1bp,clip]{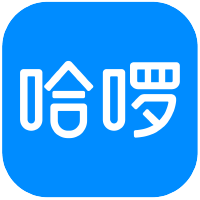}
\titlelogo[trim=32bp 86bp 56bp 96bp,clip]{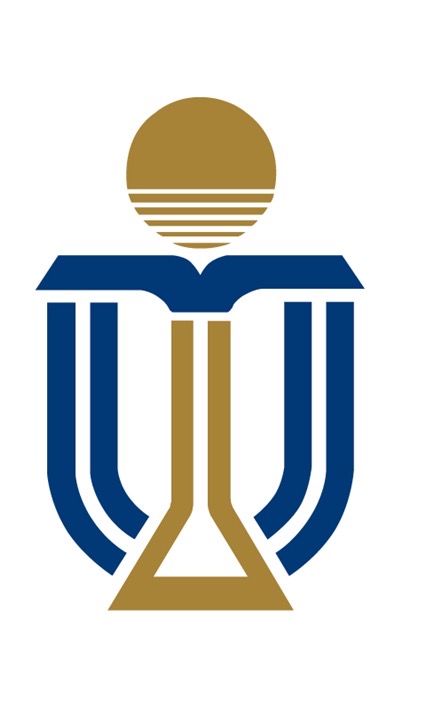}

\author[1,*]{Yixiang Yang}
\author[1]{Lizhong Zhang}
\author[2]{Xingyu Zhang}
\author[1,3]{Zhiwen Chen}
\author[2]{{Jie~Ling}}
\author[2]{Qiankun Yu}
\author[2]{Xingliang Liu}
\author[2]{Yang Lei}
\author[1,\dagger]{Zhipeng Zhang}

\affiliation[1]{AutoLab, School of Artificial Intelligence, Shanghai Jiao Tong University}
\addtolist[2]{Guangzhou Zaofu Intelligent Technology Co., Ltd.}{\affiliationlist}{\affiliationformat}{\unexpanded{\\}}
\addtolist[3]{The Hong Kong University of Science and Technology (Guangzhou)}{\affiliationlist}{\affiliationformat}{\unexpanded{\\}}

\contribution[*]{Work done at Guangzhou Zaofu Intelligent Technology Co., Ltd.}
\contribution[\dagger]{Corresponding author}

\abstract{World Action Models (WAMs) transfer scene-dynamics priors to planning by jointly forecasting future scenes and ego trajectories. However, WAMs built on video generators must generate dense future videos at inference, incurring prohibitive latency, while efficient variants discard the world branch at deployment and thus forgo test-time imagination. Even when dense video latents are replaced with patch-level DINO or JEPA features, these representations still involve long token sequences and encode decision-irrelevant information. We therefore propose SparseWAM, which forecasts in a sparse register space of only 16 tokens per frame, compressed from frozen DINOv2 features and shaped by joint planning-and-scoring pretraining to concentrate decision-relevant information. Within this space, SparseWAM employs a modality-decoupled Mixture-of-Transformers with a causal attention mask that lets action tokens condition on imagined future registers, retaining explicit imagine-then-act reasoning at deployment. Its compactness further enables low-cost multimodal sampling. As a result, our SparseWAM achieves 92.6 PDMS on NAVSIM v1, 90.1 EPDMS on NAVSIM v2, and 39.6 on NavHard at only 164 ms latency, roughly 1/5 that of DriveVA, demonstrating the efficiency of the sparse register space. Code will be released upon acceptance.}

\checkdata[Project Page]{\url{https://github.com/AutoLab-SAI-SJTU/SparseWAM}}
\begin{document}
\maketitle

\section{Introduction}
Autonomous driving is inherently interactive, requiring not only perception and understanding of the current scene but also anticipation of surrounding agents' behavior for safe and efficient planning. This requirement has motivated a growing family of World Action Models (WAMs) that jointly model future scene evolution and trajectory planning \citep{pwm, epona, eponav2, driveva, simwam, metis, drivelaw}. Because future-frame prediction provides dense, annotation-free supervision that scales with raw driving logs, these models can transfer scene-dynamics priors to planning, improving interactive reasoning and generalization, and therefore offer a compelling alternative to conventional end-to-end methods \citep{uniad, vad, transfuser, hydra-mdp, diffusiondrive} and Vision-Language-Action models \citep{emma, opendrivevla}.

Yet realizing this promise through explicit future prediction introduces a fundamental tension between imagination and efficiency. Inspired by the success of DreamZero in embodied intelligence \citep{dreamzero}, recent driving WAMs build upon pretrained video generators such as Wan2.2-5B \citep{wan}. They jointly denoise future video latents and action tokens within a unified generative process, thereby transferring the scene-dynamics priors acquired during video pretraining to trajectory planning \citep{driveva, drivewam}. While this design achieves strong performance in \textit{single-trajectory} planning, it requires future videos to be explicitly generated at inference time, resulting in substantial latency that hinders practical on-vehicle deployment. More recent approaches, including SimWAM \citep{simwam} and Metis \citep{metis}, improve efficiency by using structured attention masks to decouple action prediction from future-video generation. This allows the video-generation branch to be removed during inference. Their efficiency design, however, fundamentally changes the role of the world model. Rather than enabling the deployed policy to imagine possible futures before acting, it serves primarily as a source of representation shaping and auxiliary supervision during training.

The key may therefore lie not in abandoning test-time forecasting, but in reconsidering what the model is asked to predict. When anticipating how a traffic scene may evolve, human drivers do not reason equally about every visible region. Static background elements such as the sky and roadside buildings generally receive little attention, whereas vehicles, pedestrians, and other entities that may affect the driving decision become the focus of anticipation. Dense future-video prediction does not make this distinction. It requires the model to reconstruct the entire visual scene, devoting substantial capacity and computation to appearance details that have limited relevance to planning. As a result, the dynamics most consequential for decision-making must compete with largely irrelevant visual content for modeling capacity and learning signals.

Recent WAMs attempt to reduce this redundancy by predicting future representations alongside action trajectories in semantic latent spaces learned by models such as DINO and JEPA \citep{dinov2, vjepa2}. These representations offer more semantically meaningful prediction targets than pixels or VAE latents~\cite{wan}, but they do not fully resolve the underlying problem. Their patch-level token sequences can be comparable to, or even longer than, those of VAE latents, so forecasting them at inference time remains computationally demanding. More fundamentally, representations learned through generic visual pretraining are optimized to capture broadly useful visual semantics rather than the particular aspects of a scene that matter for driving decisions. They may therefore continue to encode background appearance and other variations that are visually salient but behaviorally irrelevant. These limitations point to a more fundamental question. \textbf{\emph{What representation space would allow a WAM to imagine the future efficiently while remaining tightly aligned with the decisions it must make?}}

\begin{wrapfigure}{r}{0.42\textwidth}
  \centering
  \includegraphics[width=\linewidth]{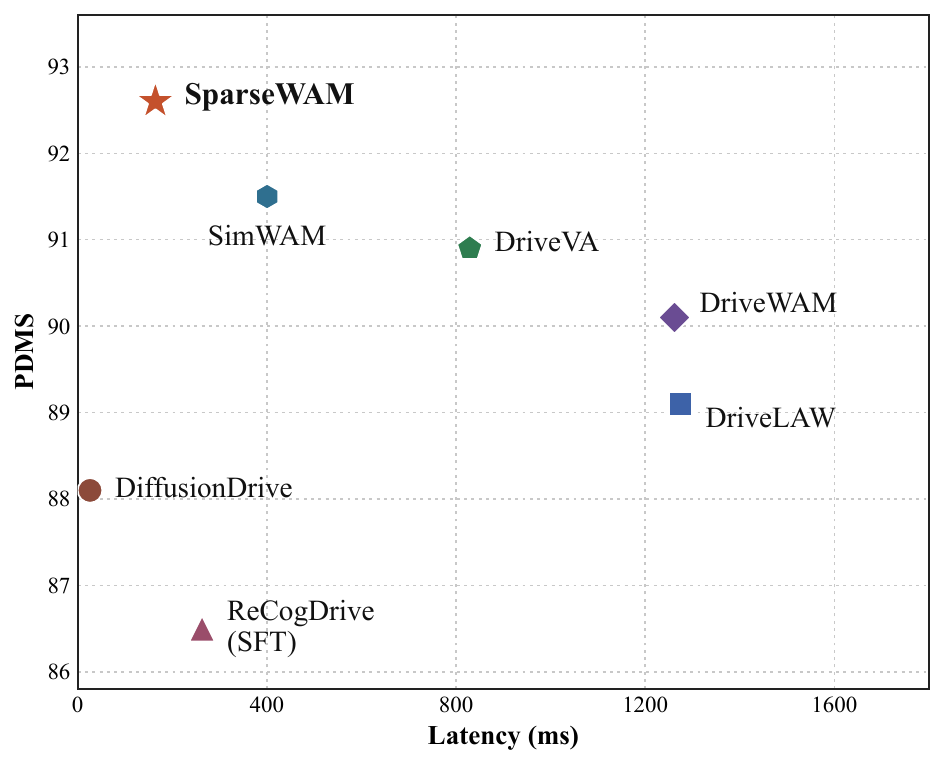}
  \caption{SparseWAM achieves the best PDMS with the lowest latency compared to WAM planners on NAVSIM.}
  \label{fig:pdms-latency}
  \vspace{4pt}
\end{wrapfigure}
We address this challenge by learning a decision-aligned sparse register space with only 16 tokens per frame. Learnable register tokens \citep{registers} compress the dense feature maps produced by a frozen DINOv2 encoder \citep{dinov2} into a compact representation, which is further shaped through joint planning-and-scoring pretraining. The planning objective encourages the registers to retain scene information that informs trajectory prediction, while the scoring objective complements it by preserving safety-critical cues that may not directly determine the nominal trajectory but remain essential for evaluating its consequences. The resulting registers therefore serve not merely as a compressed visual representation, but as a compact decision-oriented abstraction of the driving scene. We examine this property through counterfactual scene-editing probes that remove lead vehicles, insert pedestrians, or alter weather and illumination. The learned representation responds strongly to interventions that change driving decisions while remaining comparatively insensitive to appearance variations that leave those decisions unchanged. This compact representation makes it possible to preserve explicit test-time imagination without incurring the cost of dense video or feature prediction. Building on it, we introduce \textbf{SparseWAM}, a lightweight Mixture-of-Transformers denoiser \citep{mot} that uses rectified flow \citep{rectflow, flow_matching} to model future registers together with ego trajectories. Its modality-specific transformer branches reduce interference between world prediction and action generation while still allowing the two processes to exchange information. More importantly, we impose a causal attention structure that prevents future registers from attending to action tokens, while allowing action tokens to condition on denoised future registers. SparseWAM consequently learns a directed generative process in which it first anticipates how the scene may evolve and then plans in light of that imagined future. Unlike approaches that remove world prediction for efficiency, SparseWAM retains explicit imagine-then-act reasoning in the deployed policy.



This efficiency extends naturally to multimodal planning. With only 16 tokens per future frame, SparseWAM can efficiently generate diverse future–trajectory pairs by denoising multiple independent noise samples. As shown in Fig.~\ref{fig:pdms-latency}, it achieves a state-of-the-art PDMS of 92.6 with only 164 ms latency, outperforming the recent DriveVA (90.9 PDMS, 829 ms) with 1/5 inference latency. Even under single-mode inference, SparseWAM delivers performance comparable to or better than many existing single-trajectory WAMs, with a latency of just 122 ms. Rather than trading imagination for efficiency, SparseWAM makes explicit future reasoning both multimodal and practical.

Our contributions are: \textbf{(i)} We construct a driving-oriented sparse register space obtained through joint planning-and-scoring pretraining, and design counterfactual scene-editing probes that verify its sensitivity is concentrated on decision-relevant scene elements rather than appearance-related nuisance factors. \textbf{(ii)} We develop a WAM framework operating in the proposed sparse register latent space, which adopts an MoT architecture that prevents cross-modal contamination and employs a causal attention mask. \textbf{(iii)} Experiments across multiple NAVSIM benchmarks show that SparseWAM attains $92.6$ PDMS at lower latency, surpassing state-of-the-art WAMs, and remains highly competitive even under the single-mode setting, demonstrating the efficiency of the method.


\section{Related Work}

\subsection{End-to-End Autonomous Driving}
End-to-end autonomous driving has progressed along two complementary lines. Classical sensor-to-planner pipelines map multi-sensor observations to trajectories through unified differentiable stacks~\citep{transfuser, ST-P3,uniad,vad,hydra-mdp,vadv2,UAD, diffusiondrive,goalflow,flowad}, typically optimizing planning with auxiliary perception or scoring objectives. In parallel, vision--language--action (VLA) models incorporate language instructions and semantic scene understanding before producing actions~\citep{emma,DriveVLM,drivegpt4,lmdrive,Senna,omnidrive,opendrivevla}. Both paradigms have substantially advanced closed-loop planning by mapping current observations or semantic instructions directly to actions. However, these approaches mainly infer actions from current observations, leaving future scene evolution implicitly modeled. This motivates WAMs, which explicitly incorporate predictive world dynamics into end-to-end planning.
\subsection{World-Action Models for Autonomous Driving}
WAMs jointly model future scene evolution and ego actions, and their effectiveness depends heavily on the choice of future representation. Video-based WAMs inherit dynamics priors from large-scale video models by coupling future visual prediction with action generation~\citep{drivinggpt,epona,driveva,drivewam, drivelaw}. These methods demonstrate the value of predictive world dynamics, but their dense video prediction objective inevitably requires modeling substantial visual details beyond those required for driving decisions.
To reduce inference cost, SimWAM~\cite{simwam} and Metis~\cite{metis} use structured attention masks to decouple action prediction from generated futures, and DriveVLA-W0~\cite{drivevlaw0} uses future-image prediction as training-time supervision with a lightweight action expert. These designs preserve representation learning during training but forgo test-time imagination. However, these approaches mainly reduce computation by avoiding future generation, rather than reconsidering what future information is necessary for planning.
Latent-space WAMs instead forecast future scene states in a learned latent space~\cite{latentwam,worlddrive,wajepa}. These methods improve semantic abstraction, yet they still rely on visual representations that are not explicitly optimized for driving decisions. This motivates SparseWAM, which forecasts future dynamics in a driving-oriented latent space.

\subsection{Visual Representations for Autonomous Driving}
Self-supervised encoders such as DINOv2, V-JEPA, and MAE learn semantically rich features that transfer across vision tasks~\citep{dinov2,vjepa2,mae}. Driving-oriented methods further tailor such representations to the driving domain. Drive-JEPA adapts JEPA-style latent prediction to driving videos~\citep{drivejepa}, and DrivoR compresses multi-camera features into a small set of register tokens for trajectory generation and scoring~\citep{drivor}. However, these representations mainly serve as inputs to a current-timestep planner rather than as a space for forecasting future scene evolution. SparseWAM instead takes this compressed and driving-adapted space as the prediction target of future scene evolution.

\section{Preliminaries}
\label{sec:preliminary}

\paragraph{Problem Formulation.}
Given a navigation command $l_t$, historical RGB observations $I_{t-n:t}$, and ego state
$q_t\in\mathbb{R}^{d_q}$, a world action model (WAM) jointly predicts the future scene and the
ego trajectory $a_{t:t+T}\in\mathbb{R}^{T\times 3}$, where each waypoint is $(x,y,\psi)$ in the
ego frame and $T$ denotes future horizon. Previous WAMs perform this prediction in dense latent spaces, such as VAE latents
\citep{wan} or DINO/V-JEPA features \citep{dinov2, vjepa2}, with $N_p$ tokens per frame. We instead
encode each frame into sparse queries and model their evolution jointly with planning,
\begin{equation}
\label{eq:sparse-wam}
Q_{t-n:t}=\mathcal{E}\left(I_{t-n:t}\right),
\qquad
\left(a_{t+1:t+T},\,Q_{t+1:t+T}\right)\sim p_\theta\!\left(\cdot\mid Q_{t-n:t},\,q_t,\,l_t\right),
\end{equation}
where $Q_i\in\mathbb{R}^{N_r\times d}$ and $N_r=16\ll N_p$. We denote the condition by
$c=(Q_{t-n:t},q_t,l_t)$ and the future queries by $Q_f$.

\paragraph{Rectified Flow.}
We model this joint distribution with rectified flow \citep{rectflow, flow_matching}. Given a
clean sample $x$ and noise $\epsilon\sim\mathcal{N}(0,I)$, the model regresses the velocity of
the linear interpolation path,
\begin{equation}
\label{eq:fm}
x^{\sigma}=(1-\sigma)\,x+\sigma\,\epsilon,
\qquad
\mathcal{L}_{\mathrm{FM}}
=\mathbb{E}_{x,\epsilon,\sigma}
\left\lVert v_\theta\!\left(x^{\sigma},\sigma,c\right)-(\epsilon-x)\right\rVert_2^{2},
\end{equation}
where $\sigma\in[0,1]$. At inference, samples are drawn by integrating $v_\theta$ from
$\sigma=1$ to $0$ with an Euler solver. Since $v_\theta$ models the full conditional distribution, varying the
initial noise yields distinct yet plausible trajectories under the same condition $c$, naturally
supporting multi-modal planning.

\section{Method}
\label{sec:method}

\subsection{Pretraining of the Sparse Query Space}
\label{subsec:pretrain}

\textbf{Register Perceiver.}
Self-supervised latents such as DINO~\cite{dinov2} and V-JEPA~\cite{vjepa2} are semantically
richer than pixels or VAE latents~\citep{wan}, yet their token counts are no smaller. We therefore compress the semantic content
of the DINO latent space using register tokens~\cite{registers}. Concretely, given a single
frame $I\in\mathbb{R}^{3\times H_I\times W_I}$, we concatenate its patch embedding sequence
$P=[p_1,\dots,p_{N_p}]$ with a set of learnable register tokens $R=[r_1,\dots,r_{N_r}]$ and
feed the concatenation to a ViT~\cite{ViT}. After the registers have interacted with the
image tokens across all Transformer layers, we retain only the register portion of the output
sequence as the scene representation,
\begin{equation}
\label{eq:register}
\bigl[\tilde{P};\,Q\bigr]=\mathrm{ViT}\bigl([P;\,R]\bigr),
\qquad Q\in\mathbb{R}^{N_r\times d},\quad N_r=16 .
\end{equation}
Because the number of registers is decoupled from the input resolution, the scene token count
drops by roughly two orders of magnitude relative to a patch-level representation, which is
what makes the lightweight WAM architecture of Section~\ref{subsec:mot} feasible. In practice the ViT backbone uses frozen DINOv2 weights, and only
the register tokens and LoRA adapters \citep{lora} are trained, so that generic visual
priors are preserved in domain-specific adaptation to driving.

\begin{figure*}[t]
  \centering
  \includegraphics[width=\textwidth]{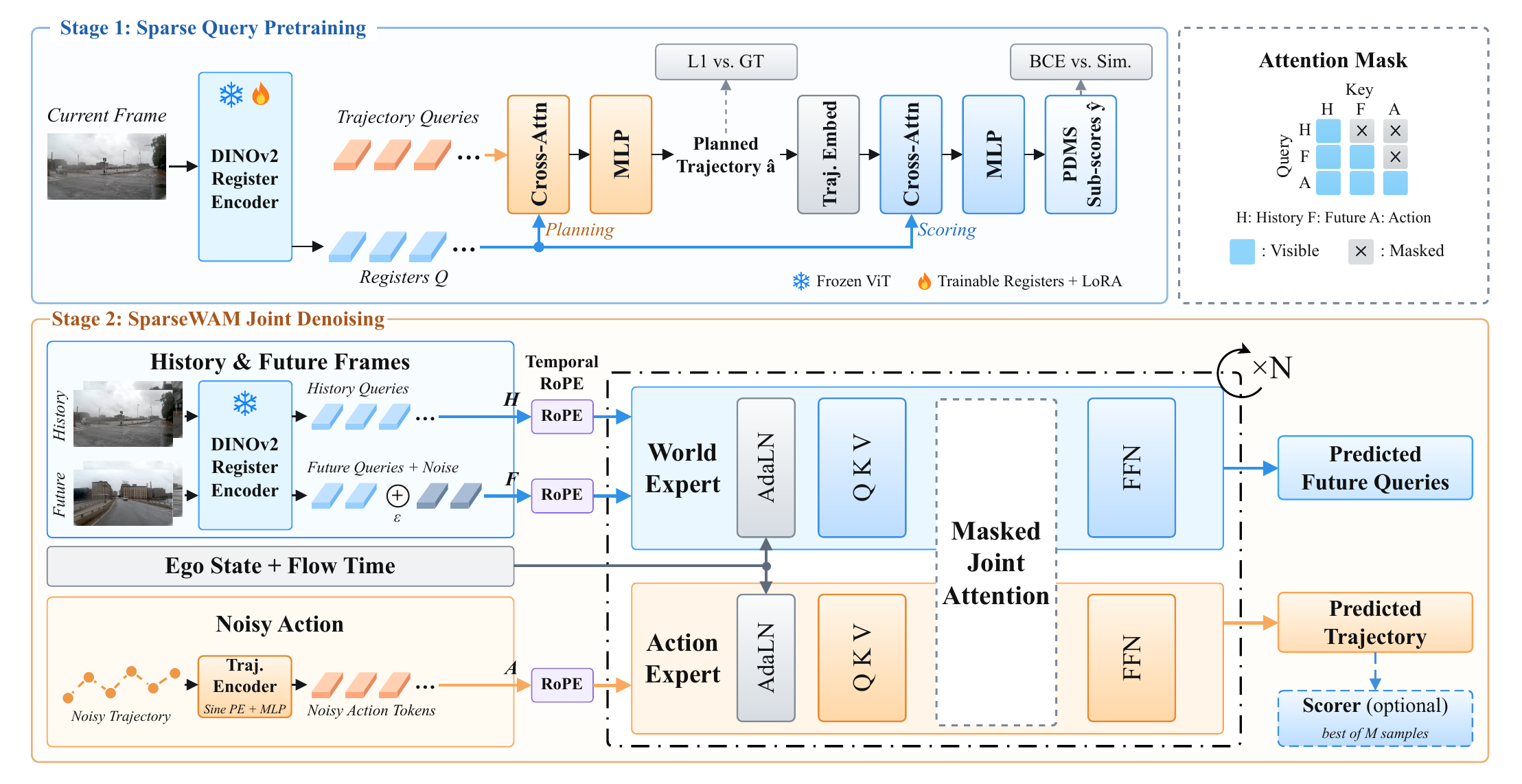}
  \caption{Overview of SparseWAM.
  Stage~1 pretrains a register encoder on a frozen DINOv2 backbone with a planning head and a scoring head.
  Stage~2 encodes history and future frames into sparse queries, perturbs the future queries and the trajectory, and jointly denoises them with a MoT under a causal attention mask.
  An optional scorer selects the best trajectory among $M$ samples.}
  \label{fig:sparsewam}
\end{figure*}

\textbf{Pretraining tasks.}
To make the registers carry sufficient scene information while attending as little as possible
to driving-irrelevant background, we shape the register encoder with two complementary tasks,
planning and scoring. As shown in Stage 1 of Fig.~\ref{fig:sparsewam}, the planning branch initializes a set of trajectory queries
$A$, lets them interact with the registers,
and decodes a planned trajectory through an MLP. The scoring branch reuses the same trajectory
representation, attends to the registers once more, and predicts the PDMS sub-metrics of that
trajectory \citep{navsim}.
The pretraining loss combines an imitation term with a score-distillation term
\citep{hydra-mdp},
\begin{equation}
\label{eq:pretrain-loss}
\mathcal{L}_{\mathrm{pre}}=\bigl\lVert \hat{a}-a^{\mathrm{gt}}\bigr\rVert_1
+\lambda_{\mathrm{sc}}\sum_{m}\mathrm{BCE}\bigl(\hat{y}_m,\,y_m^{\mathrm{sim}}\bigr),
\end{equation}
where $y^{\mathrm{sim}}$ denotes the sub-metric labels provided by the simulator. The two
objectives are complementary. The planning objective drives the registers towards scene
elements that are highly relevant to trajectory prediction, whereas the scoring objective
requires them not to omit elements that need not affect
the imitation error against the expert trajectory, yet can decisively change whether a
trajectory is feasible \citep{hydra-mdp, sparsedrivev2}. We will probe the resulting space with counterfactual scene
edits and find its sensitivity concentrated on decision-relevant scene content
rather than on background nuisance factors in Section~\ref{sec:experiments}. Once pretraining is complete, the encoder
$\mathcal{E}_{\mathrm{reg}}$ is frozen in its entirety and provides stable prediction targets
for the WAM.

\subsection{Efficient MoT Architecture}
\label{subsec:mot}

\textbf{MoT architecture.} Scene queries and trajectories are heterogeneous signals. Forcing both through a shared set of weights therefore risks contamination between the two modalities. To keep their representation spaces separate while still allowing sufficient interaction, we adopt a lightweight Mixture-of-Transformers (MoT) architecture \citep{mot} shown in Fig.~\ref{fig:sparsewam}, comprising a world expert of width $d_S$ and an action expert of width $d_A$, each with its own QKV projections, output projection, and FFN. The world expert takes a clean segment $\mathcal{H}=\{Q_{t-1},Q_t\}$ and a noisy
segment $\mathcal{F}$ (the future keyframe queries $\{Q_{t+2},Q_{t+4}\}$ linearly mixed with
Gaussian noise). The action expert reads a single segment $\mathcal{A}$. We represent the trajectory by the increments between consecutive waypoints and standardize each channel with its training-set statistics, so that displacement and heading share the scale of the Gaussian prior, and then perturb the result in the same way. Within each layer, both experts modulate their own hidden states and emit their own queries, keys, and values. These are concatenated along the token dimension, attended jointly under the causal attention mask, and split back so that each expert applies its own output projection and FFN. After the final layer, the two experts output the velocity fields $v^{S}_\theta$ and $v^{A}_\theta$.

\textbf{Condition injection and positional encoding.}
Within both experts, the noise level $\sigma$ of the respective modality, together with the ego
state and the navigation command, is injected into the DiT blocks through AdaLN \cite{dit}.
The conditioning segment $\mathcal{H}$ always takes $\sigma=0$, so that clean observations and
the futures to be denoised are distinguished explicitly at the level of conditioning. We also add a one-dimensional rotary positional embedding to $\mathcal{H}$, $\mathcal{F}$, and $\mathcal{A}$ along the same frame time axis, letting scene tokens model temporal relations
and action tokens attend to temporally nearby futures, thereby aligning future prediction with planning.

\textbf{Causal attention mask.}
The attention mask imposed on the joint attention specifies the direction of information flow
among the three segments, as illustrated in Fig.~\ref{fig:sparsewam}. The segment $\mathcal{H}$
attends only to itself, so the conditioning representation is never contaminated by the noisy
segments. $\mathcal{A}$ may
attend to partially denoised $\mathcal{F}$ within the same step, whereas $\mathcal{F}$ does
not attend to $\mathcal{A}$. This causal information flow makes the model carry out an
explicit imagine-then-act process at inference time. Unlike SimWAM and Metis
\citep{simwam, metis}, which use masks to sever the dependence of action prediction on future
prediction altogether and thereby discard the world model at deployment, our mask retains
explicit imagination at test time.

\subsection{Training Objective and Multi-modal Inference}
\label{subsec:objective}

\textbf{Training objective.}
Given the condition $c=(Q_{t-n:t},q_t,l_t)$, we supervise the
velocity field predicted for each modality with its rectified-flow target and sum the two terms,
\begin{equation}
\label{eq:loss}
\mathcal{L}=\mathbb{E}\Bigl[\bigl\lVert v^{S}_\theta-(\epsilon_S-Q_f)\bigr\rVert_2^{2}\Bigr]
+\lambda_A(\rho)\,\mathbb{E}\Bigl[\bigl\lVert v^{A}_\theta-(\epsilon_A-\tilde{a})\bigr\rVert_2^{2}\Bigr],
\end{equation}
where $v^{S}_\theta$ and $v^{A}_\theta$ are the velocities predicted by the scene and action
experts, $\epsilon_S$ and $\epsilon_A$ are independent Gaussian noises drawn for the two
branches, and $\tilde{a}$ is the normalized trajectory. The action weight $\lambda_A$ is fixed to $1$ for the first half of training and
then increases linearly to $2$ with the training progress $\rho\in[0,1]$, so that the model
first learns scene dynamics before shifting its emphasis towards trajectory accuracy.

\textbf{Multi-modal inference.}
Since flow matching models a conditional distribution rather than its mean, SparseWAM
obtains multi-modal plans by repeated sampling. At inference, $\mathcal{H}$ is encoded once,
and $M$ independent noise samples are denoised in parallel, yielding
$M$ candidate trajectories. The frozen pretraining scorer from Section~\ref{subsec:pretrain} then
selects the highest-scoring candidate \citep{wote, sparsedrivev2}. As it operates on the
current-frame registers and the ego state, it adds no image-encoding cost. Owing to the
lightweight denoiser, multi-modal inference increases the latency by only about $35\%$ relative to
single-mode inference while substantially improving planning performance
(Section~\ref{sec:experiments}).

\section{Experiments}
\label{sec:experiments}
\input{tables/navsim_v1}
\subsection{Experimental Setup}
\label{subsec:datasets}

\textbf{Datasets and Metrics.}
NAVSIM~\citep{navsim} is a planning-oriented benchmark built upon OpenScene, a $2$\,Hz redistribution of nuPlan~\citep{nuplan}. We train on the navtrain split with $103{,}288$ samples and evaluate on the navtest split with $12{,}146$ samples. 
Performance is measured by the PDM Score (PDMS), which combines five closed-loop submetrics, namely
No Collision (NC), Drivable Area Compliance (DAC), Ego Progress (EP), Time-to-Collision (TTC),
and Comfort (C).
NAVSIM v2~\citep{navsimv2} adopts a two-stage pseudo-simulation. Both stages are scored with the Extended PDM Score (EPDMS). Compared with PDMS, EPDMS adds driving direction compliance (DDC) and traffic light compliance (TLC) as penalties, and lane keeping (LK), history comfort (HC), and extended comfort (EC) as weighted terms.

\textbf{Implementation details.}
All experiments use an input resolution of $1148\times672$. The Register Perceiver uses DINOv2
ViT-S/14~\citep{dinov2} with rank-32 LoRA~\citep{lora} and is pretrained on the single front-camera view using the $103$K samples of navtrain. The WAM predicts $\{Q_{t+2},Q_{t+4}\}$ from
$\{Q_{t-1},Q_t\}$ (subscripts denote time in seconds), and $\mathcal{A}$ holds one token per
waypoint for $T=8$ waypoints over $4$\,s. Our MoT has 5 layers with $d_S=512$ and $d_A=256$,
and is trained on navtrain for 50 epochs using AdamW~\citep{adamw} with a learning rate
of $10^{-4}$ and a cosine annealing schedule with linear warm-up.
At inference, we use a 10-step Euler solver and draw $M=64$ samples for multi-modal inference.

\input{tables/navsim_v2}
\input{tables/navhard}

\subsection{Main Results}
As shown in Tab.~\ref{tab:main}, with only a single front camera, SparseWAM achieves SOTA PDMS of 92.6 on NAVSIM navtest.
It outperforms all world-model-based planners, exceeding the strongest one, DriveVA, by 1.7 PDMS.
Notably, SparseWAM improves EP by 3.1 (86.6 vs.\ 83.5) while also surpassing DriveVA in NC, indicating that imagining the future allows the model to anticipate the behavior of surrounding agents and drive more efficiently without compromising safety.
On the more comprehensive NAVSIM v2 benchmarks (Tab.~\ref{tab:navsimv2} and Tab.~\ref{tab:navhard}), SparseWAM achieves the SOTA EPDMS on both navtest and navhard, outperforming DriveLAW by 1.5 and 9.0 EPDMS, respectively.
These results show that future imagination of sparse register space remains advantageous under more comprehensive evaluations that better approximate closed-loop driving.

\subsection{Ablation Studies}

\begin{table*}[t]
  \centering
  \begin{minipage}[t]{0.48\textwidth}
    \centering
    \caption{Ablation of future prediction and MoT on NAVSIM \texttt{navtest}.}
    \label{tab:ablation-mot}
    
    \scriptsize
    \setlength{\tabcolsep}{2.6pt}
    \renewcommand{\arraystretch}{0.92}
    \setlength{\extrarowheight}{0.5pt}
    \begin{tabular}{l | c | c c c c c}
      \toprule
      Method & PDMS$\uparrow$ & NC$\uparrow$ & DAC$\uparrow$ & EP$\uparrow$ & TTC$\uparrow$ & C$\uparrow$ \\
      \midrule
      Action-only         & 87.9 & 98.3 & 96.1 & 82.1 & 94.8 & 100.0 \\
      + Future prediction & 88.5 & 98.6 & 96.4 & 82.5 & 95.3 & 100.0 \\
      \rowcolor{oursbg}
      + MoT (\textbf{ours}) & \textbf{89.1} & \textbf{98.7} & \textbf{96.7} & \textbf{83.1} & \textbf{95.6} & 100.0 \\
      \bottomrule
    \end{tabular}
  \end{minipage}\hfill
  \begin{minipage}[t]{0.50\textwidth}
    \centering
    \caption{Model size and latency on NAVSIM \texttt{navtest}. $\dagger$ means single-mode inference.}
    \label{tab:efficiency}
    \vspace{-2pt}
    \scriptsize
    \setlength{\tabcolsep}{3pt}
    \renewcommand{\arraystretch}{0.92}
    \setlength{\extrarowheight}{0.5pt}
    \begin{tabular}{l | c c | c | c}
      \toprule
      Method &
      \begin{tabular}[t]{@{}c@{}}Params\\[-1pt](train)$\downarrow$\end{tabular} &
      \begin{tabular}[t]{@{}c@{}}Params\\[-1pt](test)$\downarrow$\end{tabular} &
      PDMS$\uparrow$ &
      \begin{tabular}[t]{@{}c@{}}Latency\\[-1pt](ms)$\downarrow$\end{tabular} \\
      \midrule
      DriveVA          & 5.72B & 5.72B & 90.9 & 829.0 \\
      SimWAM (no RL)   &
      \begin{tabular}[t]{@{}c@{}}6.02B\end{tabular} &
      1.02B & 90.3 & 400.4 \\
      \midrule
      \rowcolor{oursbg}
      SparseWAM $\dagger$ & \textbf{51.8M} & \textbf{51.8M} & 89.1 & \textbf{122.0} \\
      \rowcolor{oursbg}
      SparseWAM & \textbf{51.8M} & 57.9M & \textbf{92.6} & 164.3 \\
      \bottomrule
    \end{tabular}
  \end{minipage}
\end{table*}
\paragraph{Ablation on key designs.}
To validate the contribution of each design in SparseWAM, we incrementally add components on NAVSIM navtest, with results reported in Tab.~\ref{tab:ablation-mot}. To rule out the influence of scorer selection, all variants are evaluated in the single-mode setting. The \emph{action-only} baseline, a DiT that denoises
only the trajectory conditioned on historical registers, reaches 87.9 PDMS. Next, jointly denoising
future-keyframe registers in the same DiT raises it to 88.5, with gains in NC and TTC,
indicating that modeling the future scene helps anticipate surrounding agents, thereby reducing collisions and potential collision risks. Finally, we split the shared-parameter denoiser into a world expert and an action expert following the MoT architecture, which further improves the PDMS to 89.1, indicating
modality-specific parameters reduce cross-modal interference and better promote positive
transfer from future prediction to planning.

\paragraph{Model size and inference latency.}
Tab.~\ref{tab:efficiency} compares model size and latency, measured on a single H20 GPU with
batch size 1. Because WAMs built on video generation models denoise in a dense VAE latent
space, they are large and slow: SimWAM still retains 1.02B parameters at deployment, and
DriveVA requires 829.0\,ms per inference. In contrast, SparseWAM operates on only 16 tokens
per frame, requiring 51.8M parameters and 122\,ms, which is $3.3\times$ and $6.8\times$ faster
than SimWAM and DriveVA, respectively, while still imagining the future at test time. With
multi-modal planning, SparseWAM further raises PDMS to 92.6 with only 6.1M extra parameters, at
roughly one fifth of DriveVA's latency (164.3\,ms). Although dense WAMs could in principle also perform multi-modal inference by drawing multiple
samples, their already high per-sample cost would be further multiplied, making it impractical
for deployment. In contrast, our compact register space keeps multi-sample inference cheap.
\begin{table*}[t]
  \centering
  \begin{minipage}[t]{0.50\textwidth}
    \centering
    \caption{Ablation on the attention mask between future imagination and action.}
    \label{tab:ablation_attn}
    
    \scriptsize
    \setlength{\tabcolsep}{2.4pt}
    \renewcommand{\arraystretch}{0.92}
    \setlength{\extrarowheight}{0.5pt}
    \begin{tabular}{l | c | c c c c c}
      \toprule
      Attention & PDMS$\uparrow$ & NC$\uparrow$ & DAC$\uparrow$ & EP$\uparrow$ & TTC$\uparrow$ & Comfort$\uparrow$ \\
      \midrule
      Isolated      & 88.7 & 98.4 & \textbf{96.7} & 82.8 & 95.2 & \textbf{100} \\
      Bidirectional & 88.7 & 98.5 & 96.6 & 82.8 & 95.2 & \textbf{100} \\
      \rowcolor{oursbg}
      Causal (\textbf{ours}) & \textbf{89.1} & \textbf{98.7} & \textbf{96.7} & \textbf{83.1} & \textbf{95.6} & \textbf{100} \\
      \bottomrule
    \end{tabular}
  \end{minipage}\hfill
  \begin{minipage}[t]{0.48\textwidth}
    \centering
    \caption{Performance with different visual encoders on NAVSIM \texttt{navtest}.}
    \label{tab:encoder-pdms}
    
    \scriptsize
    \setlength{\tabcolsep}{1.8pt}
    \renewcommand{\arraystretch}{0.92}
    \setlength{\extrarowheight}{0.5pt}
    \begin{tabular}{l | c | c c c c c}
      \toprule
      Method & PDMS$\uparrow$ & NC$\uparrow$ & DAC$\uparrow$ & EP$\uparrow$ & TTC$\uparrow$ & C$\uparrow$ \\
      \midrule
      DINOv2         & 87.8 & 98.2 & 96.2 & 82.4 & 94.1 & 100.0 \\
      V-JEPA 2.1     & 87.8 & 98.1 & 96.3 & 82.3 & 94.2 & 100.0 \\
      \midrule
      \rowcolor{oursbg}
      SparseWAM (\textbf{ours}) & \textbf{89.1} & \textbf{98.7} & \textbf{96.7} & \textbf{83.1} & \textbf{95.6} & 100.0 \\
      \bottomrule
    \end{tabular}
  \end{minipage}
\end{table*}
\paragraph{Ablation on the causal mask.}
As shown in Tab.~\ref{tab:ablation_attn}, the \textit{Isolated} mask adopted by SimWAM and Metis~\citep{simwam, metis} blocks action tokens from attending to future registers, so future prediction serves only as a training-time auxiliary objective (88.7 PDMS).
The \textit{Bidirectional} mask lets actions read the imagined future, but also exposes it to noisy action tokens, which corrupts the imagination and yields no gain (88.7 PDMS).
The \textit{Causal} mask removes this interference while keeping the future accessible to actions, achieving the best PDMS of 89.1 and indicating that the performance gain stems from the imagine-then-act process.

\subsection{Comparison and Visualization of Representations}

\paragraph{Representation ablation and visualization.}
We replace the register space in our architecture with alternative representation spaces while keeping the MoT architecture unchanged (more details in appendix). As shown in Tab.~\ref{tab:encoder-pdms}, the WAMs built on DINOv2 and V-JEPA 2.1 both get PDMS of 87.8, yet fall below the single-sample performance of the register-based variant trained with the same recipe. This further demonstrates that our representation space is better suited for WAM prediction and yields stronger positive transfer to planning. We further visualize the register-to-image attention maps in the appendix, which show that the registers concentrate on planning-critical elements such as lead vehicles and road signs while largely ignoring the background, explaining why such a compact space suffices for future prediction and planning.

\paragraph{Element sensitivity of representations.}
To further evaluate how sensitive different representation spaces are to scene-level background versus planning-relevant objects, we adopt a counterfactual image editing protocol. Using rule-based
filtering on 3D ground-truth boxes, we first mine 400 daytime \emph{car-following} frames and 400
daytime \emph{open-road} frames (no object ahead) from navtest. We then edit these base images with FLUX.2-klein-4B~\cite{flux} under four categories, which are two planning-relevant edits $\mathcal{P}$ (removing the lead vehicle in car-following scenes and inserting a pedestrian in open-road scenes), and two background edits $\mathcal{B}$ (converting the whole image to nighttime and to rainy weather). Each category contains 400 pairs of original and edited images. Then, each pair $(x, x')$ is fed into the frozen encoder under evaluation to obtain flattened representation vectors $z(x)$ and $z(x')$. Intuitively, a larger representation shift under a given edit type indicates that the representation is more sensitive to that type of element.
\paragraph{Metric.}
The sensitivity $S$ of each pair is defined as the relative $L_2$ displacement computed over the entire flattened vector:
\begin{equation}
S(x, x') = \frac{\lVert z(x') - z(x) \rVert_2}{\lVert z(x) \rVert_2 + \epsilon}.
\end{equation}

\begin{wraptable}{r}{0.5\textwidth}
  \centering
  \vspace{-12pt}
  \caption{Token budget and counterfactual sensitivity of scene representations.}
  \label{tab:fbsr}
  
  \scriptsize
  \setlength{\tabcolsep}{3.2pt}
  \renewcommand{\arraystretch}{0.92}
  \setlength{\extrarowheight}{0.5pt}
  \begin{tabular}{l | c | c  c | c}
    \toprule
    Encoder & Tokens$\downarrow$ & $S_{\mathrm{plan}}\!\uparrow$ & $S_{\mathrm{bg}}\!\downarrow$ & FBSR$\uparrow$ \\
    \midrule
    DINOv2      & 3936 & 0.297 & 0.692 & 0.43 \\
    V-JEPA 2.1  & 3024 & 0.189 & 0.551 & 0.34 \\
    Wan2.2 VAE  & 756  & 0.206 & 1.061 & 0.19 \\
    \midrule
    \rowcolor{oursbg}
    Sparse Register (\textbf{ours}) & \textbf{16} & \textbf{0.736} & \textbf{0.520} & \textbf{1.41} \\
    \bottomrule
  \end{tabular}
  \vspace{-8pt}
\end{wraptable}
Averaging $S$ over the pairs in $\mathcal{P}$ and $\mathcal{B}$ yields $S_{\text{plan}}$ and
$S_{\text{bg}}$, and we define the foreground-to-background sensitivity ratio
$\text{FBSR}=S_{\text{plan}}/S_{\text{bg}}$.
A higher FBSR indicates that the
representation is more sensitive to planning-relevant objects relative to appearance changes. As shown in Tab.~\ref{tab:fbsr}, we find that the Sparse Register space is the only representation whose FBSR exceeds 1, demonstrating that it responds more strongly to planning-relevant object changes than to lighting and weather changes, whereas all other representations are dominated by appearance variations. We also find that DINO and V-JEPA achieve higher FBSR than the VAE encoder of Wan-2.2 owing to their richer semantic content.

\subsection{Qualitative Results}
\begin{wrapfigure}{r}{0.40\linewidth}
    \centering
    \vspace{-2\baselineskip}
    \includegraphics[width=\linewidth]{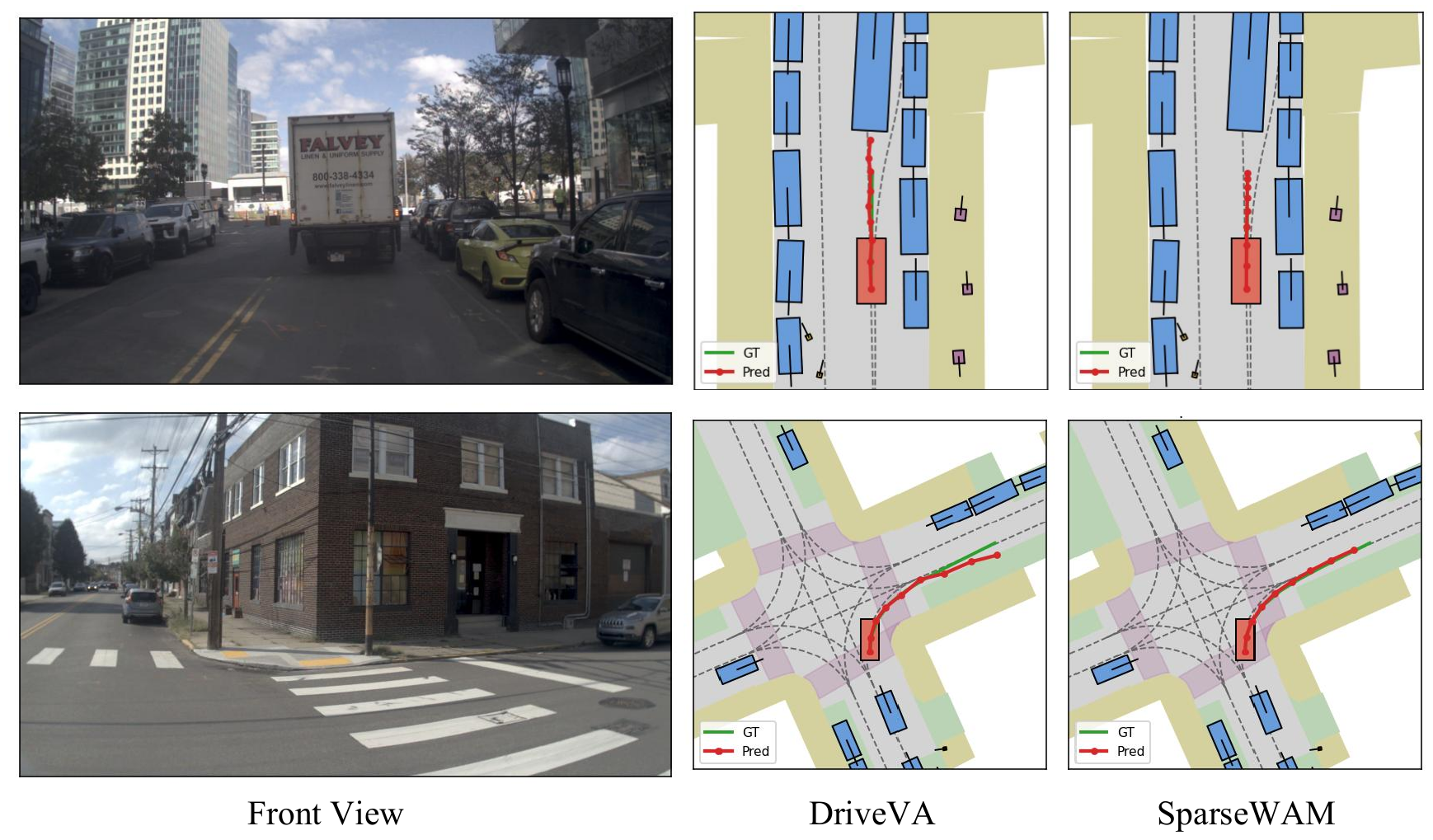}
    \caption{Qualitative comparison between DriveVA and SparseWAM.}
    \label{fig:qualitative}
\end{wrapfigure}
As shown in Fig.~\ref{fig:qualitative}, we compare the planning results of SparseWAM and DriveVA~\cite{driveva} in two representative scenarios. The first row shows a car-following scenario with a large truck ahead of the ego vehicle. SparseWAM attends to the motion of the leading vehicle and plans a decelerating trajectory, braking in time to avoid a rear-end collision. DriveVA plans too aggressively, and its trajectory extends into the leading vehicle, which results in a collision. The second row shows a left turn at an intersection. SparseWAM produces a smooth and accurate turning trajectory that stays within the drivable area. DriveVA's turning trajectory hits the curb and leaves the drivable area.

\section{Conclusion}
\label{sec:conclusion}
We presented SparseWAM, a world action model that imagines the future in a decision-aligned sparse
register space. Built by compressing a frozen
DINOv2 encoder into 16 registers per frame and shaped by joint planning-and-scoring pretraining,
this space is markedly more sensitive to planning-relevant objects relative to appearance changes.
On top of it, a lightweight Mixture-of-Transformers with a causal attention mask jointly denoises
future registers and trajectories, retaining explicit imagine-then-act reasoning at deployment.
Our SparseWAM achieves SOTA PDMS of 92.6 on NAVSIM at 164\,ms while remaining competitive on NAVSIM v2, showing that decision-aligned sparsity is a possible
practical path towards efficient WAMs. Extending it to multi-view inputs and closed-loop evaluation
is left for future work.
\clearpage

\bibliographystyle{plainnat}
\bibliography{main}

\appendix
\section{Appendix}

\subsection{Details of Counterfactual Scene Editing}
\label{app:edit}

\paragraph{Base-frame mining.}
All base frames are front-camera images from NAVSIM \texttt{navtest}. We first keep daytime
frames and then apply rule-based filtering on the 3D ground-truth boxes in the ego frame.
A frame is labeled \emph{car-following} if a vehicle lies in the ego lane
($|y|<1.5$\,m) with a longitudinal distance of $5$--$30$\,m, and
\emph{open-road} if no annotated agent lies within $40$\,m ahead and $3$\,m
laterally. We sample 400 frames of each type.

\begin{figure}[h]
  \centering
  \begin{minipage}{\linewidth}
    \centering
    \includegraphics[width=\linewidth]{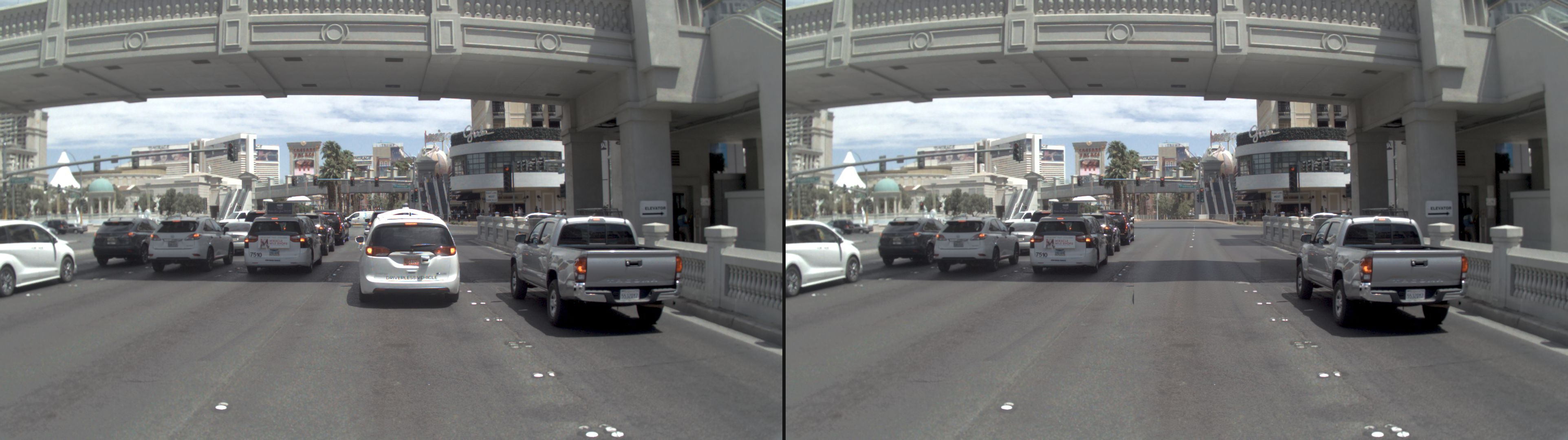}\\[-1pt]
    {\small (a) $\mathcal{P}$: remove lead vehicle}\\[4pt]
    \includegraphics[width=\linewidth]{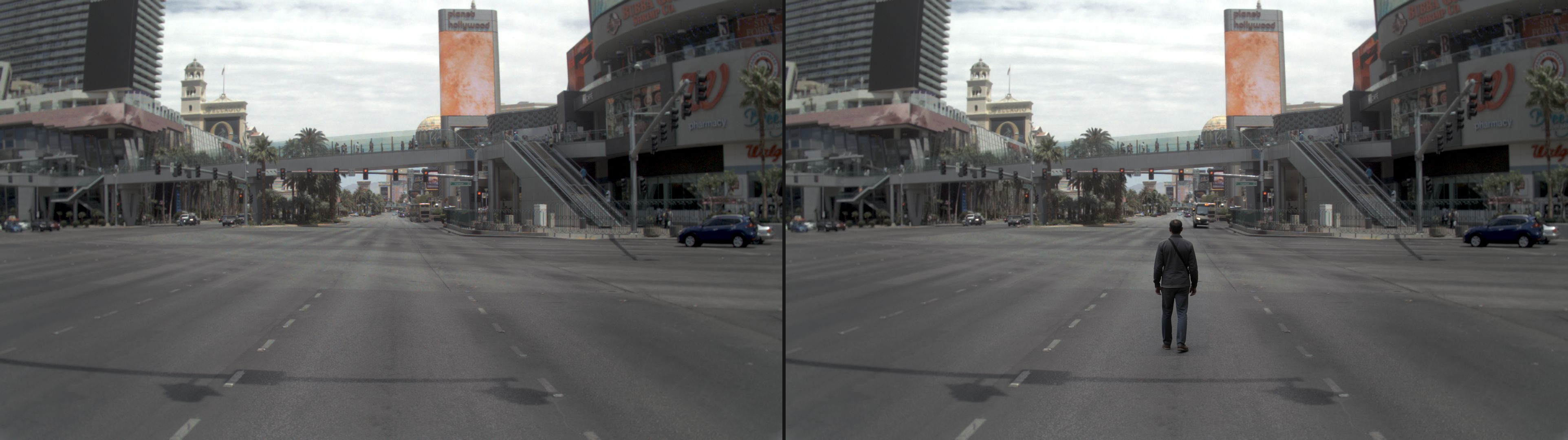}\\[-1pt]
    {\small (b) $\mathcal{P}$: insert pedestrian}\\[4pt]
    \includegraphics[width=\linewidth]{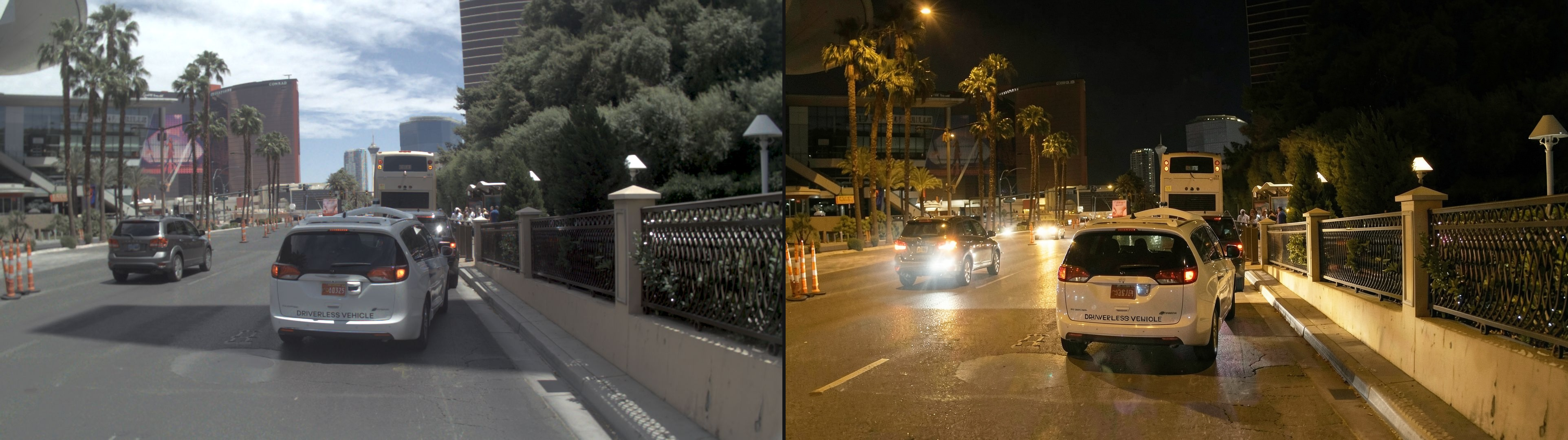}\\[-1pt]
    {\small (c) $\mathcal{B}$: nighttime}\\[4pt]
    \includegraphics[width=\linewidth]{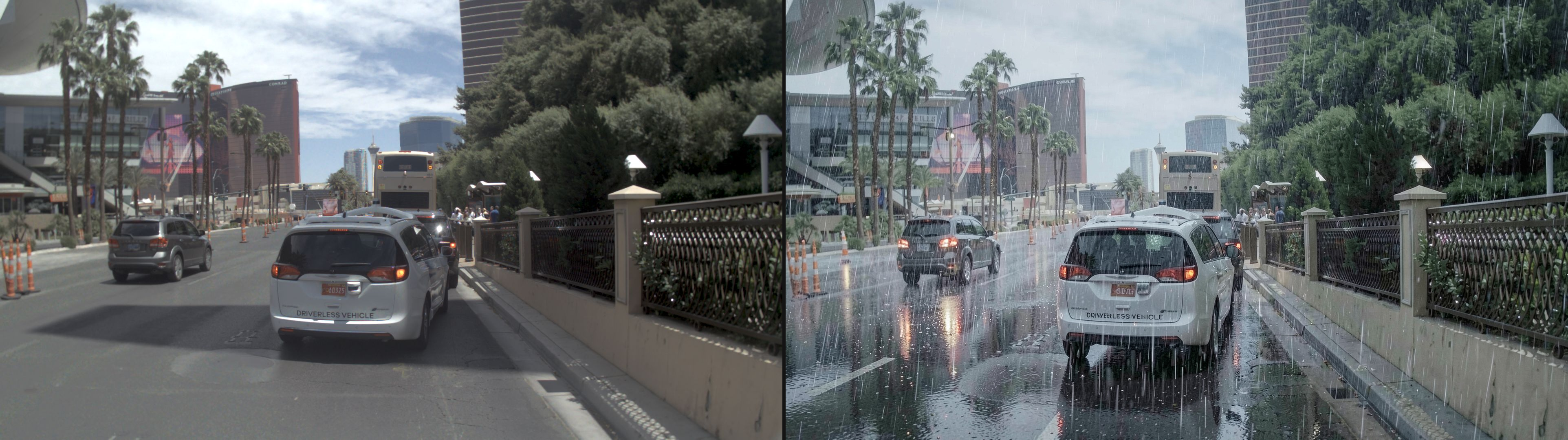}\\[-1pt]
    {\small (d) $\mathcal{B}$: rainy weather}
  \end{minipage}
  \caption{Examples of counterfactual edits. In each pair, the left image is the original
  \texttt{navtest} frame and the right image is the edited one. Planning-relevant edits
  $\mathcal{P}$ change a decision-critical object while keeping the rest of the scene intact,
  whereas background edits $\mathcal{B}$ drastically alter the appearance while keeping all
  objects and the layout unchanged.}
  \label{fig:edit-examples}
\end{figure}

\paragraph{Editing protocol.}
Each base image is edited with FLUX.2-klein-4B~\cite{flux} at its original resolution using the
instructions in Tab.~\ref{tab:edit-prompts}. Car-following frames receive the
\emph{remove-lead-vehicle} edit, open-road frames receive the \emph{insert-pedestrian} edit, and
the background edits (\emph{night} and \emph{rain}) are applied to car-following frames, so
that each of the four categories contains 400 original--edited pairs. Examples are shown in
Fig.~\ref{fig:edit-examples}.

\begin{table}[h]
  \centering
  \caption{Editing instructions for the four counterfactual categories.}
  \label{tab:edit-prompts}
  \small
  \begin{tabularx}{\linewidth}{@{}l l X@{}}
    \toprule
    Type & Category & Instruction \\
    \midrule
    $\mathcal{P}$ & Remove lead vehicle & ``Remove the vehicle directly in front of the camera and
      fill the region with the road surface, keeping everything else unchanged.'' \\
    $\mathcal{P}$ & Insert pedestrian & ``Add a pedestrian walking on the road in front of the
      camera, keeping everything else unchanged.'' \\
    $\mathcal{B}$ & Night & ``Convert the scene to nighttime with street lights on, keeping
      the layout and all objects unchanged.'' \\
    $\mathcal{B}$ & Rain & ``Convert the scene to heavy rain with a wet road surface, keeping
      the layout and all objects unchanged.'' \\
    \bottomrule
  \end{tabularx}
\end{table}

\paragraph{Human review.}
Generative editing can fail silently, e.g., by leaving the target partly visible, altering
unrelated agents or the road layout, or introducing visible artifacts. To ensure that each pair
differs \emph{only} in the intended factor, every edited image was manually inspected by
two annotators against three criteria: (i) the target edit is fully realized;
(ii) the road geometry, other agents, and camera viewpoint remain unchanged; and (iii) the
result is photorealistic without noticeable artifacts. Pairs failing any criterion were
discarded and regenerated with a new random seed, or replaced by another base frame, until
each category contained 400 valid pairs. Overall, 1.82\% of the initial edits were
rejected.

\paragraph{Feature extraction.}
All encoders are frozen and receive the same resized input. We use the last-layer patch
tokens of DINOv2 ViT-S/14 without the CLS and register tokens ($82\times48=3936$ tokens), the
image-mode patch tokens of V-JEPA 2.1 ViT-B with the input width resized to $1152$
($72\times42=3024$ tokens), the latent of the Wan2.2 VAE ($756$ tokens), and the $16$ output registers of
our Register Perceiver. Each representation is flattened into a single vector before computing
the sensitivity.

\begin{figure}[h]
  \centering
  \includegraphics[width=0.9\linewidth]{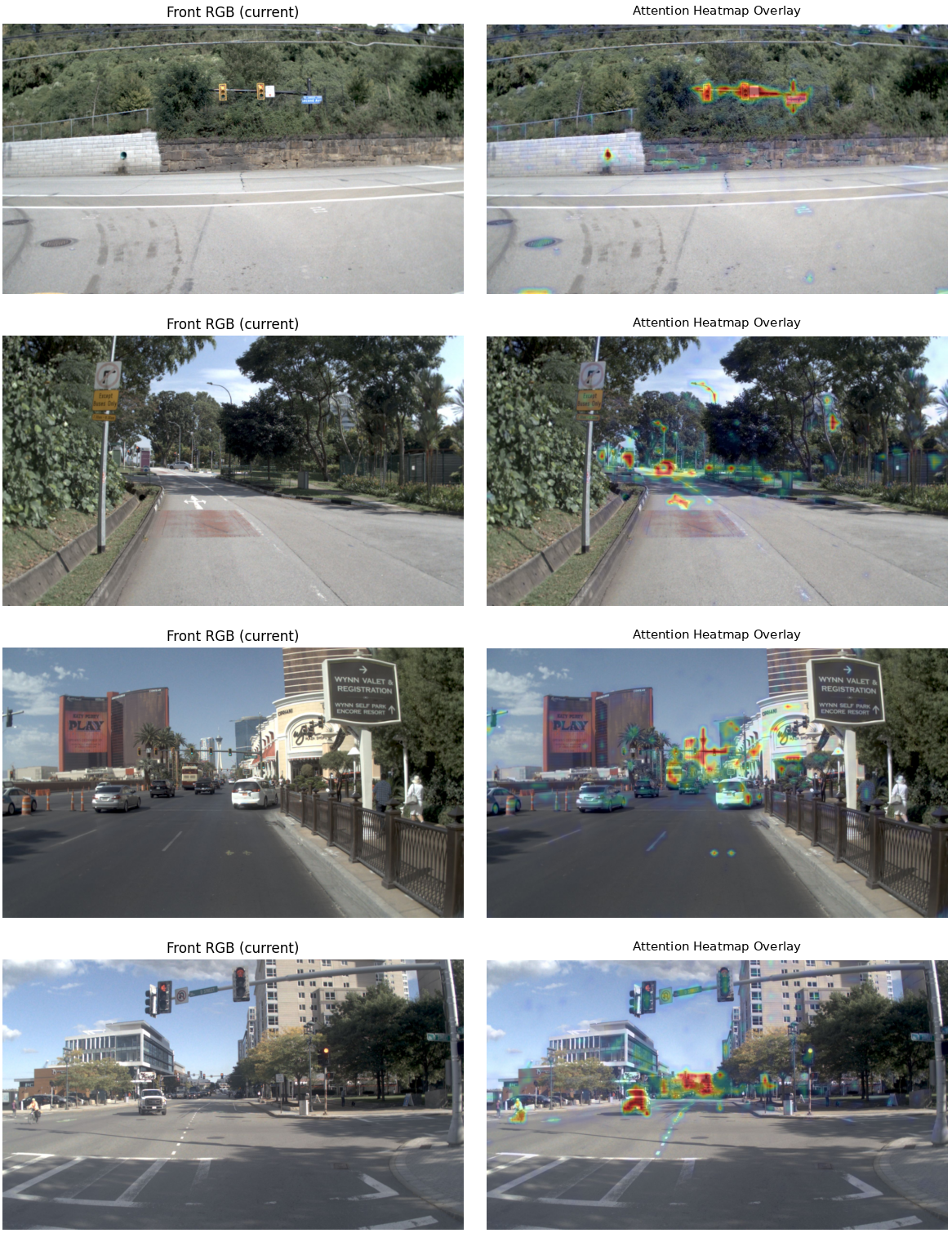}
  \caption{Register-to-image attention in the last DINOv2 block. Left: input front-camera
  frame. Right: attention averaged over registers and overlaid on the input frame.}
  \label{fig:register-attn}
\end{figure}

\subsection{Visualization of Register Attention}
\label{app:register-attn}

To see which image regions the registers absorb, we take the attention weights from the
register tokens to all patch tokens in the last block of the DINOv2 backbone, average them over
heads and registers, and reshape them to the $82\times48$ patch grid. The map is bilinearly
upsampled to the input resolution and overlaid on the RGB image, with low-attention regions
left transparent.

As shown in Fig.~\ref{fig:register-attn}, the attention is sparse and concentrates on
planning-critical elements rather than on salient but driving-irrelevant content. In the first
row, it focuses almost exclusively on the distant traffic lights and road signs, while the
large wall and vegetation receive little attention. In the second row, it spreads along the
road ahead and the markings on the road surface. In the third row, it concentrates on the
vehicles ahead in the ego and adjacent lanes, while ignoring the large billboard and the
roadside sign despite their high visual saliency. In the fourth row, it highlights the lead
vehicle, the cyclist on the left, and the traffic lights, while the sky and the building
facades are largely ignored. These observations are consistent with the high FBSR of the
register space in Tab.~\ref{tab:fbsr} and explain why only $16$ tokens per frame suffice for
future prediction and planning.

\subsection{Implementation of Dense Representation Baselines}
\label{app:dense-baselines}

For the representation ablation in Tab.~\ref{tab:encoder-pdms}, we replace the learnable
register queries of the Register Perceiver with a frozen off-the-shelf visual encoder followed
by spatial pooling, and train the same MoT world--action model on top of it. The two variants
differ only in the encoder; the WAM architecture and training recipe are shared.

\paragraph{Scene tokenization.}
Each front-camera frame at $1148\times672$ is encoded independently; history and future frames
share the same encoder.
\begin{itemize}
  \item \emph{DINOv2}: the official ViT-S/14~\citep{dinov2} with patch size $14$. We discard the
  CLS token and its $4$ register tokens and keep the last-layer patch features, yielding a dense
  map of $384\times48\times82$ ($C\times H\times W$).
  \item \emph{V-JEPA 2.1}: the ViT-B encoder~\citep{vjepa2} in image mode, i.e., a single frame
  ($T=1$) through the image patch embedding. The input width is bilinearly resized from $1148$
  to $1152$ to be divisible by the patch size $16$, yielding a dense map of
  $768\times42\times72$.
\end{itemize}
The dense map is compressed by \texttt{AdaptiveAvgPool2d} to a $10\times10$ grid, i.e., $100$
tokens per frame, and projected to $256$ dimensions by a linear neck. Both the encoder and the
neck are kept frozen in evaluation mode throughout training, so that the WAM predicts a fixed
target space, analogous to the frozen $\mathcal{E}_{\mathrm{reg}}$ in SparseWAM.

\paragraph{Joint sequence and attention.}
The joint sequence is $[\mathcal{H}\,|\,\mathcal{F}\,|\,\mathcal{A}]$. The history segment
$\mathcal{H}$ contains the clean tokens of the frames at $-1$\,s and $0$\,s ($200$ tokens) with
the flow time fixed to $0$; the future segment $\mathcal{F}$ contains the noisy tokens of the
frames at $+2$\,s and $+4$\,s ($200$ tokens); and the action segment $\mathcal{A}$ contains
$8$ noisy waypoints. The attention mask is asymmetric: $\mathcal{H}$ attends only to
$\mathcal{H}$, $\mathcal{F}$ attends to $\mathcal{H}$ and $\mathcal{F}$ but not to actions, and
$\mathcal{A}$ attends to all tokens. Temporal positions are encoded with a shared 1D
RoPE whose coordinate is the frame index at $2$\,Hz; the $k$-th waypoint is
assigned index $k{+}1$, so the $+2$\,s future frame is aligned with the fourth waypoint.
Spatial positions within the $10\times10$ grid use fixed 2D sinusoidal embeddings.

\paragraph{MoT and training.}
The MoT consists of a world expert of width $512$ and an action expert of width $256$, whose
tokens are concatenated for joint attention with $12$ heads of dimension $64$. Both branches are
trained with rectified flow: a shared uniform sample $u$ is drawn per example and mapped to the
flow time of each branch by a separate SNR shift. The loss is the MSE between the predicted and
target velocities on $\mathcal{F}$ and $\mathcal{A}$ only, where a Beta-distributed timestep
weighting is applied to the action term. The action loss weight is fixed to $1$ for the first
half of training and increases linearly to $2$ afterwards, as in SparseWAM. At inference, both
segments start from pure Gaussian noise and are integrated with a $10$-step Euler solver, and
only the trajectory is output. For a fair comparison, all variants in
Tab.~\ref{tab:encoder-pdms} are evaluated with single-sample inference without the scorer.

\end{document}

%% file: tables/navsim_v1.tex
\begin{table*}[t]
  \caption{\textbf{Comparison with state-of-the-art planners on the NAVSIM \texttt{navtest} benchmark.}
  C denotes camera and L denotes LiDAR.
  The best learned result in each column is shown in \textbf{bold}.}
  \label{tab:main}
  \vspace{4pt}
  \centering
  \scriptsize
  \setlength{\tabcolsep}{4.5pt}
  \renewcommand{\arraystretch}{0.92}
  \setlength{\extrarowheight}{0.6pt}
  \setlength{\aboverulesep}{0.35ex}
  \setlength{\belowrulesep}{0.45ex}
  \begin{tabularx}{\dimexpr\linewidth-2\tabcolsep-4pt\relax}{
    >{\raggedright\arraybackslash\hsize=3.6\hsize}X
    >{\raggedright\arraybackslash\hsize=1.05\hsize}X
    >{\centering\arraybackslash\hsize=0.75\hsize}X
    *{6}{>{\centering\arraybackslash\hsize=0.6\hsize}X}
  }
    \toprule
    Method & Reference & Sensors & NC$\uparrow$ & DAC$\uparrow$ & EP$\uparrow$ & TTC$\uparrow$ & C$\uparrow$ & PDMS$\uparrow$ \\
    \midrule
    Human Agent & -- & -- & 100.0 & 100.0 & 87.5 & 100.0 & 99.9 & 94.8 \\
    \midrule
    \rowcolor{grpbg}
    \multicolumn{9}{c}{\textit{Traditional E2E planners}} \\
    UniAD~\cite{uniad}             & CVPR'23    & 6$\times$C   & 97.8 & 91.9 & 78.8 & 92.9 & 100.0 & 83.4 \\
    TransFuser~\cite{transfuser}    & TPAMI'22   & 3$\times$C+L & 97.7 & 92.8 & 79.2 & 92.8 & 100.0 & 84.0 \\
    DiffusionDrive~\cite{diffusiondrive}  & CVPR'25    & 3$\times$C+L & 98.2 & 96.2 & 82.2 & 94.7 & 100.0 & 88.1 \\
    WoTE~\cite{wote}              & ICCV'25    & 3$\times$C+L & 98.5 & 96.8 & 81.9 & 94.9 & 99.9 & 88.3 \\
    DiffusionDriveV2~\cite{diffusiondrivev2} & arXiv'25   & 3$\times$C+L & 98.3 & 97.9 & 87.5 & 94.8 & 99.9 & 91.2 \\
    SparseDriveV2~\cite{sparsedrivev2}    & arXiv'26   & 3$\times$C   & 98.5 & 98.4 & \textbf{88.6} & 95.0 & 99.9 & 92.0 \\
    \rowcolor{grpbg}
    \multicolumn{9}{c}{\textit{VLM-based planners}} \\
    OneDrive~\cite{onedrive}        & arXiv'26   & 1$\times$C & 98.4 & 95.2 & 81.1 & 94.9 & 100.0 & 86.8 \\
    ReCogDrive~\cite{recogdrive}       & ICLR'26    & 1$\times$C & 97.9 & 97.3 & 87.3 & 94.9 & 100.0 & 90.8 \\
    AutoVLA~\cite{autovla}        & NeurIPS'25 & 3$\times$C & 98.4 & 95.6 & 81.9 & 98.0 & 99.9 & 89.1 \\
    DriveVLA-W0~\cite{drivevlaw0}      & ICLR'26    & 1$\times$C & 98.7 & \textbf{99.1} & 83.3 & 95.3 & 99.3 & 90.2 \\
    SGDrive~\cite{sgdrive}          & CVPR'26    & 1$\times$C & 98.6 & 97.8 & 85.8 & 96.2 & 100.0 & 91.1 \\
    ExploreVLA~\cite{explorevla}     & ECCV'26    & 1$\times$C & 98.8 & 98.4 & 83.5 & 96.5 & 99.9 & 90.4 \\
    \rowcolor{grpbg}
    \multicolumn{9}{c}{\textit{World-model-based planners}} \\
    Epona~\cite{epona}         & ICCV'25    & 1$\times$C & 97.9 & 95.1 & 80.4 & 93.8 & 99.9 & 86.2 \\
    PWM~\cite{pwm}            & NeurIPS'25 & 1$\times$C & 98.6 & 95.9 & 81.8 & 95.4 & 100.0 & 88.1 \\
    DriveLaW~\cite{drivelaw}         & CVPR'26    & 1$\times$C & 99.0 & 97.1 & 81.3 & 96.7 & 100.0 & 89.1 \\
    DriveVA~\cite{driveva}          & ECCV'26    & 1$\times$C & 99.2 & 97.5 & 83.5 & \textbf{98.7} & 100.0 & 90.9 \\
    \midrule
    \rowcolor{oursbg}
    SparseWAM (\textbf{ours}) & -- & 1$\times$C & \textbf{99.3} & 98.8 & 86.6 & 97.5 & 100.0 & \textbf{92.6} \\
    \bottomrule
  \end{tabularx}
\end{table*}

%% file: tables/navsim_v2.tex
\begin{table*}[t]
  \caption{\textbf{Performance on the NAVSIM-v2 \texttt{navtest} benchmark.}
  * indicates training with reinforcement learning.
  The best learned result in each column is shown in \textbf{bold}.}
  \label{tab:navsimv2}
  \vspace{4pt}
  \centering
  \scriptsize
  \setlength{\tabcolsep}{1.6pt}
  \renewcommand{\arraystretch}{0.92}
  \setlength{\extrarowheight}{0.6pt}
  \setlength{\aboverulesep}{0.35ex}
  \setlength{\belowrulesep}{0.45ex}
  \begin{tabularx}{\linewidth}{
    l
    >{\centering\arraybackslash}X
    *{10}{>{\hspace{1pt}}c<{\hspace{1pt}}}
  }
    \toprule
    Method & Reference & NC$\uparrow$ & DAC$\uparrow$ & DDC$\uparrow$ & TLC$\uparrow$ & EP$\uparrow$ & TTC$\uparrow$ & LK$\uparrow$ & HC$\uparrow$ & EC$\uparrow$ & EPDMS$\uparrow$ \\
    \midrule
    Human Agent & -- & 100.0 & 100.0 & 99.8 & 100.0 & 87.4 & 100.0 & 100.0 & 98.1 & 90.1 & 90.3 \\
    \midrule
    \rowcolor{grpbg}
    \multicolumn{12}{c}{\textit{Traditional E2E planners}} \\
    TransFuser~\cite{transfuser}   & TPAMI'22 & 96.9 & 89.9 & 97.8 & 99.7 & 87.1 & 95.4 & 92.7 & 98.3 & 87.2 & 76.7 \\
    DiffusionDrive~\cite{diffusiondrive} & CVPR'25  & 98.2 & 95.9 & 99.4 & 99.8 & 87.5 & 97.3 & 96.8 & 98.3 & \textbf{87.7} & 84.5 \\
    \rowcolor{grpbg}
    \multicolumn{12}{c}{\textit{VLM-based planners}} \\
    ReCogDrive*~\cite{recogdrive}      & ICLR'26  & 98.3 & 95.2 & 99.5 & 99.8 & 87.1 & 97.5 & 96.6 & 98.3 & 86.5 & 83.6 \\
    SGDrive~\cite{sgdrive}          & CVPR'26  & 98.6 & 94.3 & 99.5 & \textbf{99.9} & 86.0 & 97.9 & 96.1 & 98.3 & 85.9 & 86.2 \\
    DriveFine*~\cite{drivefine}      & arXiv'26 & 98.7 & 97.3 & 99.5 & 99.8 & 88.7 & 97.8 & \textbf{97.7} & \textbf{98.4} & 83.8 & 89.7 \\
    \rowcolor{grpbg}
    \multicolumn{12}{c}{\textit{World-model-based planners}} \\
    DriveVLA-W0~\cite{drivevlaw0}      & ICLR'26  & 98.5 & \textbf{99.1} & 98.0 & 99.7 & 86.4 & 98.1 & 93.2 & 97.9 & 58.9 & 86.1 \\
    DriveLaW~\cite{drivelaw}         & CVPR'26  & 98.7 & 96.9 & \textbf{99.6} & 99.8 & 87.5 & 98.3 & 97.6 & \textbf{98.4} & 77.4 & 88.6 \\
    \midrule
    \rowcolor{oursbg}
    SparseWAM (\textbf{ours}) & -- & \textbf{98.8} & 98.8 & 99.5 & 99.7 & \textbf{89.0} & \textbf{98.8} & 96.8 & 98.3 & 72.8 & \textbf{90.1} \\
    \bottomrule
  \end{tabularx}
  \vspace{-5pt}
\end{table*}

%% file: tables/navhard.tex
\begin{table*}[t]
  \caption{\textbf{Performance on the NAVSIM-v2 \texttt{navhard} benchmark.}
  S1/S2 denote the per-stage results of the official two-stage protocol, and EPDMS is the overall score.
  The best learned result in each column is shown in \textbf{bold}.}
  \label{tab:navhard}
  \vspace{4pt}
  \centering
  \scriptsize
  \setlength{\tabcolsep}{1.6pt}
  \renewcommand{\arraystretch}{0.92}
  \setlength{\extrarowheight}{0.6pt}
  \setlength{\aboverulesep}{0.35ex}
  \setlength{\belowrulesep}{0.45ex}
  \begin{tabularx}{\dimexpr\linewidth-2\tabcolsep-4pt\relax}{
    >{\raggedright\arraybackslash}X
    >{\centering\arraybackslash}X
    c
    *{10}{>{\hspace{1pt}}c<{\hspace{1pt}}}
  }
    \toprule
    Method & Reference & Stage & NC$\uparrow$ & DAC$\uparrow$ & DDC$\uparrow$ & TLC$\uparrow$ & EP$\uparrow$ & TTC$\uparrow$ & LK$\uparrow$ & HC$\uparrow$ & EC$\uparrow$ & EPDMS$\uparrow$ \\
    \midrule
    \rowcolor{grpbg}
    \multicolumn{13}{c}{\textit{Traditional E2E planners}} \\
    & & S1 & 96.2 & 79.5 & 99.1 & 99.5 & 84.1 & 95.1 & 94.2 & 97.5 & 79.1 & \\
    \multirow{-2}{*}{TransFuser~\cite{transfuser}} & \multirow{-2}{*}{TPAMI'22} & S2 & 77.7 & 70.2 & 84.2 & 98.0 & 85.1 & 75.6 & 45.4 & 95.7 & \textbf{75.9} & \multirow{-2}{*}{23.1} \\
    & & S1 & 96.8 & 86.0 & 98.8 & 99.3 & 84.0 & 95.8 & 96.7 & 97.6 & \textbf{79.6} & \\
    \multirow{-2}{*}{DiffusionDrive~\cite{diffusiondrive}} & \multirow{-2}{*}{CVPR'25} & S2 & 80.1 & 72.8 & 84.4 & 98.4 & 85.9 & 76.6 & 46.4 & 96.3 & 72.8 & \multirow{-2}{*}{27.5} \\
    \rowcolor{grpbg}
    \multicolumn{13}{c}{\textit{VLM-based planners}} \\
    & & S1 & 95.8 & 87.6 & 97.8 & 99.8 & 84.4 & 94.7 & 92.9 & \textbf{97.8} & 28.9 & \\
    \multirow{-2}{*}{SGDrive~\cite{sgdrive}} & \multirow{-2}{*}{CVPR'26} & S2 & 79.4 & 65.4 & 79.1 & 98.9 & 88.9 & 75.3 & 42.7 & 96.4 & 29.6 & \multirow{-2}{*}{25.5} \\
    & & S1 & 96.4 & 78.9 & 98.7 & 99.8 & 82.6 & 95.6 & 94.4 & 97.6 & 74.2 & \\
    \multirow{-2}{*}{ReCogDrive~\cite{recogdrive}} & \multirow{-2}{*}{ICLR'26} & S2 & 80.2 & 65.0 & 82.4 & 98.7 & 85.2 & 76.9 & 43.8 & 96.6 & 71.8 & \multirow{-2}{*}{25.7} \\
    & & S1 & 97.6 & 90.0 & 99.1 & 99.3 & \textbf{84.9} & 96.7 & \textbf{97.3} & 97.6 & 72.0 & \\
    \multirow{-2}{*}{DriveFine~\cite{drivefine}} & \multirow{-2}{*}{arXiv'26} & S2 & 82.1 & 71.3 & 84.8 & 98.4 & 88.1 & 74.3 & \textbf{47.2} & \textbf{96.8} & 72.8 & \multirow{-2}{*}{30.5} \\
    \rowcolor{grpbg}
    \multicolumn{13}{c}{\textit{World-model-based planners}} \\
    & & S1 & 96.8 & 83.3 & 99.0 & 99.6 & 84.6 & 95.3 & 96.4 & 97.6 & 78.2 & \\
    \multirow{-2}{*}{DriveVLA-W0~\cite{drivevlaw0}} & \multirow{-2}{*}{ICLR'26} & S2 & 76.8 & 64.3 & 79.9 & 98.3 & \textbf{89.2} & 75.0 & 46.8 & 95.8 & 53.1 & \multirow{-2}{*}{24.4} \\
    & & S1 & 97.3 & 89.1 & \textbf{99.2} & 99.6 & 84.3 & 97.1 & 96.2 & \textbf{97.8} & 67.6 & \\
    \multirow{-2}{*}{DriveLaW~\cite{drivelaw}} & \multirow{-2}{*}{CVPR'26} & S2 & 82.5 & 67.6 & 83.5 & 98.1 & 84.8 & 78.5 & 45.8 & 96.4 & 57.3 & \multirow{-2}{*}{30.6} \\
    \midrule
    \rowcolor{oursbg}
    & & S1 & \textbf{98.1} & \textbf{94.9} & 99.1 & \textbf{100.0} & 83.0 & \textbf{98.0} & 96.0 & \textbf{97.8} & 52.4 & \\
    \rowcolor{oursbg}
    \multirow{-2}{*}{SparseWAM (\textbf{ours})} & \multirow{-2}{*}{--} & S2 & \textbf{83.0} & \textbf{84.0} & \textbf{87.1} & \textbf{99.2} & 84.4 & \textbf{79.1} & 45.9 & 96.4 & 64.9 & \multirow{-2}{*}{\textbf{39.6}} \\
    \bottomrule
  \end{tabularx}
\end{table*}

%% file: paper.bbl
\begin{thebibliography}{58}
\providecommand{\natexlab}[1]{#1}
\providecommand{\url}[1]{\texttt{#1}}
\expandafter\ifx\csname urlstyle\endcsname\relax
  \providecommand{\doi}[1]{doi: #1}\else
  \providecommand{\doi}{doi: \begingroup \urlstyle{rm}\Url}\fi

\bibitem[Assran et~al.(2025)Assran, Bardes, Fan, Garrido, Howes, Mojtaba, Komeili, Muckley, Rizvi, Roberts, Sinha, Zholus, Arnaud, Gejji, Martin, Hogan, Dugas, Bojanowski, Khalidov, Labatut, Massa, Szafraniec, Krishnakumar, Li, Ma, Chandar, Meier, LeCun, Rabbat, and Ballas]{vjepa2}
Mido Assran, Adrien Bardes, David Fan, Quentin Garrido, Russell Howes, Mojtaba, Komeili, Matthew Muckley, Ammar Rizvi, Claire Roberts, Koustuv Sinha, Artem Zholus, Sergio Arnaud, Abha Gejji, Ada Martin, Francois~Robert Hogan, Daniel Dugas, Piotr Bojanowski, Vasil Khalidov, Patrick Labatut, Francisco Massa, Marc Szafraniec, Kapil Krishnakumar, Yong Li, Xiaodong Ma, Sarath Chandar, Franziska Meier, Yann LeCun, Michael Rabbat, and Nicolas Ballas.
\newblock V-jepa 2: Self-supervised video models enable understanding, prediction and planning, 2025.
\newblock URL \url{https://arxiv.org/abs/2506.09985}.

\bibitem[Caesar et~al.(2021)Caesar, Kabzan, Tan, Fong, Wolff, Lang, Fletcher, Beijbom, and Omari]{nuplan}
Holger Caesar, Juraj Kabzan, Kok~Seang Tan, Whye~Kit Fong, Eric Wolff, Alex Lang, Luke Fletcher, Oscar Beijbom, and Sammy Omari.
\newblock nuplan: A closed-loop ml-based planning benchmark for autonomous vehicles.
\newblock \emph{arXiv preprint arXiv:2106.11810}, 2021.

\bibitem[Cao et~al.(2026)Cao, Hallgarten, Li, Dauner, Gu, Wang, Miron, Aiello, Li, Gilitschenski, Ivanovic, Pavone, Geiger, and Chitta]{navsimv2}
Wei Cao, Marcel Hallgarten, Tianyu Li, Daniel Dauner, Xunjiang Gu, Caojun Wang, Yakov Miron, Marco Aiello, Hongyang Li, Igor Gilitschenski, Boris Ivanovic, Marco Pavone, Andreas Geiger, and Kashyap Chitta.
\newblock Pseudo-simulation for autonomous driving, 2026.
\newblock URL \url{https://arxiv.org/abs/2506.04218}.

\bibitem[Chen et~al.(2024{\natexlab{a}})Chen, Jiang, Gao, Liao, Xu, Zhang, Huang, Liu, and Wang]{vadv2}
Shaoyu Chen, Bo~Jiang, Hao Gao, Bencheng Liao, Qing Xu, Qian Zhang, Chang Huang, Wenyu Liu, and Xinggang Wang.
\newblock Vadv2: End-to-end vectorized autonomous driving via probabilistic planning.
\newblock \emph{arXiv preprint arXiv:2402.13243}, 2024{\natexlab{a}}.

\bibitem[Chen et~al.(2024{\natexlab{b}})Chen, Wang, and Zhang]{drivinggpt}
Yuntao Chen, Yuqi Wang, and Zhaoxiang Zhang.
\newblock Drivinggpt: Unifying driving world modeling and planning with multi-modal autoregressive transformers.
\newblock \emph{arXiv preprint arXiv:2412.18607}, 2024{\natexlab{b}}.

\bibitem[Dang et~al.(2026)Dang, Ang, Li, Tian, Wang, Li, Ye, Ma, Chen, and Wang]{drivefine}
Chenxu Dang, Sining Ang, Yongkang Li, Haochen Tian, Jie Wang, Guang Li, Hangjun Ye, Jie Ma, Long Chen, and Yan Wang.
\newblock Drivefine: Refining-augmented masked diffusion vla for precise and robust driving, 2026.
\newblock URL \url{https://arxiv.org/abs/2602.14577}.

\bibitem[Darcet et~al.(2024)Darcet, Oquab, Mairal, and Bojanowski]{registers}
Timoth{\'e}e Darcet, Maxime Oquab, Julien Mairal, and Piotr Bojanowski.
\newblock Vision transformers need registers.
\newblock In \emph{International conference on learning representations}, volume 2024, pages 2632--2652, 2024.

\bibitem[Dauner et~al.(2024)Dauner, Hallgarten, Li, Weng, Huang, Yang, Li, Gilitschenski, Ivanovic, Pavone, Geiger, and Chitta]{navsim}
Daniel Dauner, Marcel Hallgarten, Tianyu Li, Xinshuo Weng, Zhiyu Huang, Zetong Yang, Hongyang Li, Igor Gilitschenski, Boris Ivanovic, Marco Pavone, Andreas Geiger, and Kashyap Chitta.
\newblock Navsim: Data-driven non-reactive autonomous vehicle simulation and benchmarking.
\newblock In \emph{Advances in Neural Information Processing Systems (NeurIPS)}, 2024.

\bibitem[Dosovitskiy et~al.(2020)Dosovitskiy, Beyer, Kolesnikov, Weissenborn, Zhai, Unterthiner, Dehghani, Minderer, Heigold, Gelly, et~al.]{ViT}
Alexey Dosovitskiy, Lucas Beyer, Alexander Kolesnikov, Dirk Weissenborn, Xiaohua Zhai, Thomas Unterthiner, Mostafa Dehghani, Matthias Minderer, Georg Heigold, Sylvain Gelly, et~al.
\newblock An image is worth 16x16 words: Transformers for image recognition at scale.
\newblock \emph{arXiv preprint arXiv:2010.11929}, 2020.

\bibitem[Gui et~al.(2026)Gui, Zhang, Yan, Han, Gong, Tan, zhong Xu, and Shen]{worlddrive}
Xingtai Gui, Meijie Zhang, Tianyi Yan, Wencheng Han, Jiahao Gong, Feiyang Tan, Cheng zhong Xu, and Jianbing Shen.
\newblock Bridging scene generation and planning: Driving with world model via unifying vision and motion representation, 2026.
\newblock URL \url{https://arxiv.org/abs/2603.14948}.

\bibitem[Guo et~al.(2024)Guo, Zhang, He, Wang, and Jing]{UAD}
Mingzhe Guo, Zhipeng Zhang, Yuan He, Ke~Wang, and Liping Jing.
\newblock End-to-end autonomous driving without costly modularization and 3d manual annotation.
\newblock \emph{arXiv preprint arXiv:2406.17680}, 2024.

\bibitem[Guo et~al.(2026)Guo, Yang, Han, Zhang, Li, Wan, and Zhang]{flowad}
Mingzhe Guo, Yixiang Yang, Chuanrong Han, Rufeng Zhang, Shirui Li, Ji~Wan, and Zhipeng Zhang.
\newblock Flowad: Ego-scene interactive modeling for autonomous driving.
\newblock \emph{arXiv preprint arXiv:2603.13399}, 2026.

\bibitem[He et~al.(2021)He, Chen, Xie, Li, Dollár, and Girshick]{mae}
Kaiming He, Xinlei Chen, Saining Xie, Yanghao Li, Piotr Dollár, and Ross Girshick.
\newblock Masked autoencoders are scalable vision learners, 2021.
\newblock URL \url{https://arxiv.org/abs/2111.06377}.

\bibitem[Hu et~al.(2021)Hu, Shen, Wallis, Allen-Zhu, Li, Wang, Wang, and Chen]{lora}
Edward~J Hu, Yelong Shen, Phillip Wallis, Zeyuan Allen-Zhu, Yuanzhi Li, Shean Wang, Lu~Wang, and Weizhu Chen.
\newblock Lora: Low-rank adaptation of large language models.
\newblock \emph{arXiv preprint arXiv:2106.09685}, 2021.

\bibitem[Hu et~al.(2022)Hu, Chen, Wu, Li, Yan, and Tao]{ST-P3}
Shengchao Hu, Li~Chen, Penghao Wu, Hongyang Li, Junchi Yan, and Dacheng Tao.
\newblock St-p3: End-to-end vision-based autonomous driving via spatial-temporal feature learning.
\newblock In \emph{ECCV}. Springer, 2022.

\bibitem[Hu et~al.(2023)Hu, Yang, Chen, Li, Sima, Zhu, Chai, Du, Lin, Wang, et~al.]{uniad}
Yihan Hu, Jiazhi Yang, Li~Chen, Keyu Li, Chonghao Sima, Xizhou Zhu, Siqi Chai, Senyao Du, Tianwei Lin, Wenhai Wang, et~al.
\newblock Planning-oriented autonomous driving.
\newblock In \emph{CVPR}, 2023.

\bibitem[Hwang et~al.(2024)Hwang, Xu, Lin, Hung, Ji, Choi, Huang, He, Covington, Sapp, et~al.]{emma}
Jyh-Jing Hwang, Runsheng Xu, Hubert Lin, Wei-Chih Hung, Jingwei Ji, Kristy Choi, Di~Huang, Tong He, Paul Covington, Benjamin Sapp, et~al.
\newblock Emma: End-to-end multimodal model for autonomous driving.
\newblock \emph{arXiv preprint arXiv:2410.23262}, 2024.

\bibitem[Jiang et~al.(2023)Jiang, Chen, Xu, Liao, Chen, Zhou, Zhang, Liu, Huang, and Wang]{vad}
Bo~Jiang, Shaoyu Chen, Qing Xu, Bencheng Liao, Jiajie Chen, Helong Zhou, Qian Zhang, Wenyu Liu, Chang Huang, and Xinggang Wang.
\newblock Vad: Vectorized scene representation for efficient autonomous driving.
\newblock \emph{arXiv preprint arXiv:2303.12077}, 2023.

\bibitem[Jiang et~al.(2024)Jiang, Chen, Liao, Zhang, Yin, Zhang, Huang, Liu, and Wang]{Senna}
Bo~Jiang, Shaoyu Chen, Bencheng Liao, Xingyu Zhang, Wei Yin, Qian Zhang, Chang Huang, Wenyu Liu, and Xinggang Wang.
\newblock Senna: Bridging large vision-language models and end-to-end autonomous driving.
\newblock \emph{arXiv preprint arXiv:2410.22313}, 2024.

\bibitem[Kirby et~al.(2026)Kirby, Boulch, Xu, Yin, Puy, Éloi Zablocki, Bursuc, Gidaris, Marlet, Bartoccioni, Cao, Samet, VU, and Cord]{drivor}
Ellington Kirby, Alexandre Boulch, Yihong Xu, Yuan Yin, Gilles Puy, Éloi Zablocki, Andrei Bursuc, Spyros Gidaris, Renaud Marlet, Florent Bartoccioni, Anh-Quan Cao, Nermin Samet, Tuan-Hung VU, and Matthieu Cord.
\newblock Driving on registers, 2026.
\newblock URL \url{https://arxiv.org/abs/2601.05083}.

\bibitem[Labs(2025)]{flux}
Black~Forest Labs.
\newblock {FLUX.2: Frontier Visual Intelligence}.
\newblock \url{https://bfl.ai/blog/flux-2}, 2025.

\bibitem[Li et~al.(2026{\natexlab{a}})Li, Liu, Hu, Wu, Ma, Wu, Han, Hao, Liu, Zhan, et~al.]{metis}
Jingyu Li, Zhe Liu, Dongnan Hu, Junjie Wu, Zipei Ma, Wenxiao Wu, Chao Han, Zhihui Hao, Zhikang Liu, Kun Zhan, et~al.
\newblock Metis: A generalizable and efficient world-action model for autonomous driving and urban navigation.
\newblock \emph{arXiv preprint arXiv:2606.15869}, 2026{\natexlab{a}}.

\bibitem[Li et~al.(2026{\natexlab{b}})Li, Wu, Hu, Huang, Sun, Hao, Lang, Zhu, and Zhang]{sgdrive}
Jingyu Li, Junjie Wu, Dongnan Hu, Xiangkai Huang, Bin Sun, Zhihui Hao, Xianpeng Lang, Xiatian Zhu, and Li~Zhang.
\newblock Sgdrive: Scene-to-goal hierarchical world cognition for autonomous driving, 2026{\natexlab{b}}.
\newblock URL \url{https://arxiv.org/abs/2601.05640}.

\bibitem[Li et~al.(2025{\natexlab{a}})Li, Shang, Liu, Zhan, Wang, Wang, Chen, Wang, An, Tang, Hou, Fan, and Zhang]{drivevlaw0}
Yingyan Li, Shuyao Shang, Weisong Liu, Bing Zhan, Haochen Wang, Yuqi Wang, Yuntao Chen, Xiaoman Wang, Yasong An, Chufeng Tang, Lu~Hou, Lue Fan, and Zhaoxiang Zhang.
\newblock Drivevla-w0: World models amplify data scaling law in autonomous driving, 2025{\natexlab{a}}.
\newblock URL \url{https://arxiv.org/abs/2510.12796}.

\bibitem[Li et~al.(2025{\natexlab{b}})Li, Wang, Liu, He, Fan, and Zhang]{wote}
Yingyan Li, Yuqi Wang, Yang Liu, Jiawei He, Lue Fan, and Zhaoxiang Zhang.
\newblock End-to-end driving with online trajectory evaluation via bev world model.
\newblock In \emph{Proceedings of the IEEE/CVF International Conference on Computer Vision}, pages 27137--27146, 2025{\natexlab{b}}.

\bibitem[Li et~al.(2025{\natexlab{c}})Li, Xiong, Guo, Li, Yan, Xu, Zhou, Chen, Sun, Wang, Ma, Chen, Ye, Liu, and Wang]{recogdrive}
Yongkang Li, Kaixin Xiong, Xiangyu Guo, Fang Li, Sixu Yan, Gangwei Xu, Lijun Zhou, Long Chen, Haiyang Sun, Bing Wang, Kun Ma, Guang Chen, Hangjun Ye, Wenyu Liu, and Xinggang Wang.
\newblock Recogdrive: A reinforced cognitive framework for end-to-end autonomous driving, 2025{\natexlab{c}}.
\newblock URL \url{https://arxiv.org/abs/2506.08052}.

\bibitem[Li et~al.(2024)Li, Li, Wang, Lan, Yu, Ji, Li, Zhu, Kautz, Wu, et~al.]{hydra-mdp}
Zhenxin Li, Kailin Li, Shihao Wang, Shiyi Lan, Zhiding Yu, Yishen Ji, Zhiqi Li, Ziyue Zhu, Jan Kautz, Zuxuan Wu, et~al.
\newblock Hydra-mdp: End-to-end multimodal planning with multi-target hydra-distillation.
\newblock \emph{arXiv preprint arXiv:2406.06978}, 2024.

\bibitem[Liang et~al.(2024)Liang, Yu, Luo, Iyer, Dong, Zhou, Ghosh, Lewis, Yih, Zettlemoyer, et~al.]{mot}
Weixin Liang, Lili Yu, Liang Luo, Srinivasan Iyer, Ning Dong, Chunting Zhou, Gargi Ghosh, Mike Lewis, Wen-tau Yih, Luke Zettlemoyer, et~al.
\newblock Mixture-of-transformers: A sparse and scalable architecture for multi-modal foundation models.
\newblock \emph{arXiv preprint arXiv:2411.04996}, 2024.

\bibitem[Liao et~al.(2025)Liao, Chen, Yin, Jiang, Wang, Yan, Zhang, Li, Zhang, Zhang, et~al.]{diffusiondrive}
Bencheng Liao, Shaoyu Chen, Haoran Yin, Bo~Jiang, Cheng Wang, Sixu Yan, Xinbang Zhang, Xiangyu Li, Ying Zhang, Qian Zhang, et~al.
\newblock Diffusiondrive: Truncated diffusion model for end-to-end autonomous driving.
\newblock In \emph{Proceedings of the Computer Vision and Pattern Recognition Conference}, pages 12037--12047, 2025.

\bibitem[Lipman et~al.(2022)Lipman, Chen, Ben-Hamu, Nickel, and Le]{flow_matching}
Yaron Lipman, Ricky~TQ Chen, Heli Ben-Hamu, Maximilian Nickel, and Matt Le.
\newblock Flow matching for generative modeling.
\newblock \emph{arXiv preprint arXiv:2210.02747}, 2022.

\bibitem[Liu et~al.(2026)Liu, Zhang, Liu, Cui, Xie, Chen, Ye, Yang, Nex, and Cheng]{driveva}
Mengmeng Liu, Diankun Zhang, Jiuming Liu, Jianfeng Cui, Hongwei Xie, Guang Chen, Hangjun Ye, Michael~Ying Yang, Francesco Nex, and Hao Cheng.
\newblock Driveva: Video action models are zero-shot drivers.
\newblock In \emph{European Conference on Computer Vision}, pages 315--335. Springer, 2026.

\bibitem[Liu et~al.(2022)Liu, Gong, and Liu]{rectflow}
Xingchao Liu, Chengyue Gong, and Qiang Liu.
\newblock Flow straight and fast: Learning to generate and transfer data with rectified flow.
\newblock \emph{arXiv preprint arXiv:2209.03003}, 2022.

\bibitem[Loshchilov(2017)]{adamw}
I~Loshchilov.
\newblock Decoupled weight decay regularization.
\newblock \emph{arXiv preprint arXiv:1711.05101}, 2017.

\bibitem[Oquab et~al.(2024)Oquab, Darcet, Moutakanni, Vo, Szafraniec, Khalidov, Fernandez, Haziza, Massa, El-Nouby, Assran, Ballas, Galuba, Howes, Huang, Li, Misra, Rabbat, Sharma, Synnaeve, Xu, Jegou, Mairal, Labatut, Joulin, and Bojanowski]{dinov2}
Maxime Oquab, Timothée Darcet, Théo Moutakanni, Huy Vo, Marc Szafraniec, Vasil Khalidov, Pierre Fernandez, Daniel Haziza, Francisco Massa, Alaaeldin El-Nouby, Mahmoud Assran, Nicolas Ballas, Wojciech Galuba, Russell Howes, Po-Yao Huang, Shang-Wen Li, Ishan Misra, Michael Rabbat, Vasu Sharma, Gabriel Synnaeve, Hu~Xu, Hervé Jegou, Julien Mairal, Patrick Labatut, Armand Joulin, and Piotr Bojanowski.
\newblock Dinov2: Learning robust visual features without supervision, 2024.
\newblock URL \url{https://arxiv.org/abs/2304.07193}.

\bibitem[Peebles and Xie(2023)]{dit}
William Peebles and Saining Xie.
\newblock Scalable diffusion models with transformers.
\newblock In \emph{2023 IEEE/CVF International Conference on Computer Vision (ICCV)}, pages 4172--4182. IEEE, 2023.

\bibitem[Prakash et~al.(2021)Prakash, Chitta, and Geiger]{transfuser}
Aditya Prakash, Kashyap Chitta, and Andreas Geiger.
\newblock Multi-modal fusion transformer for end-to-end autonomous driving.
\newblock In \emph{CVPR}, 2021.

\bibitem[Shao et~al.(2024)Shao, Hu, Wang, Song, Waslander, Liu, and Li]{lmdrive}
Hao Shao, Yuxuan Hu, Letian Wang, Guanglu Song, Steven~L Waslander, Yu~Liu, and Hongsheng Li.
\newblock Lmdrive: Closed-loop end-to-end driving with large language models.
\newblock In \emph{Proceedings of the IEEE/CVF Conference on Computer Vision and Pattern Recognition}, pages 15120--15130, 2024.

\bibitem[Sheng et~al.(2026)Sheng, Ye, Luo, Chen, and Ren]{explorevla}
Zihao Sheng, Xin Ye, Jingru Luo, Sikai Chen, and Liu Ren.
\newblock Explorevla: Dense world modeling and exploration for end-to-end autonomous driving.
\newblock In \emph{European Conference on Computer Vision}, pages 266--284. Springer, 2026.

\bibitem[Shi et~al.(2026)Shi, Xu, Shi, Sheng, Zhang, and Jiang]{drivewam}
Chen Shi, Jinrui Xu, Shaoshuai Shi, Kehua Sheng, Bo~Zhang, and Li~Jiang.
\newblock Drivewam: Video generative priors enable scalable world-action modeling for autonomous driving.
\newblock \emph{arXiv preprint arXiv:2605.28544}, 2026.

\bibitem[Sun et~al.(2026)Sun, Lin, Chen, Pei, Li, Shi, and Zheng]{sparsedrivev2}
Wenchao Sun, Xuewu Lin, Keyu Chen, Zixiang Pei, Xiang Li, Yining Shi, and Sifa Zheng.
\newblock Sparsedrivev2: Scoring is all you need for end-to-end autonomous driving.
\newblock \emph{arXiv preprint arXiv:2603.29163}, 2026.

\bibitem[Tian et~al.(2024)Tian, Gu, Li, Liu, Hu, Wang, Zhan, Jia, Lang, and Zhao]{DriveVLM}
Xiaoyu Tian, Junru Gu, Bailin Li, Yicheng Liu, Chenxu Hu, Yang Wang, Kun Zhan, Peng Jia, Xianpeng Lang, and Hang Zhao.
\newblock Drivevlm: The convergence of autonomous driving and large vision-language models.
\newblock \emph{arXiv preprint arXiv:2402.12289}, 2024.

\bibitem[Wan et~al.(2025)Wan, Wang, Ai, Wen, Mao, Xie, Chen, Yu, Zhao, Yang, Zeng, Wang, Zhang, Zhou, Wang, Chen, Zhu, Zhao, Yan, Huang, Feng, Zhang, Li, Wu, Chu, Feng, Zhang, Sun, Fang, Wang, Gui, Weng, Shen, Lin, Wang, Wang, Zhou, Wang, Shen, Yu, Shi, Huang, Xu, Kou, Lv, Li, Liu, Wang, Zhang, Huang, Li, Wu, Liu, Pan, Zheng, Hong, Shi, Feng, Jiang, Han, Wu, and Liu]{wan}
Team Wan, Ang Wang, Baole Ai, Bin Wen, Chaojie Mao, Chen-Wei Xie, Di~Chen, Feiwu Yu, Haiming Zhao, Jianxiao Yang, Jianyuan Zeng, Jiayu Wang, Jingfeng Zhang, Jingren Zhou, Jinkai Wang, Jixuan Chen, Kai Zhu, Kang Zhao, Keyu Yan, Lianghua Huang, Mengyang Feng, Ningyi Zhang, Pandeng Li, Pingyu Wu, Ruihang Chu, Ruili Feng, Shiwei Zhang, Siyang Sun, Tao Fang, Tianxing Wang, Tianyi Gui, Tingyu Weng, Tong Shen, Wei Lin, Wei Wang, Wei Wang, Wenmeng Zhou, Wente Wang, Wenting Shen, Wenyuan Yu, Xianzhong Shi, Xiaoming Huang, Xin Xu, Yan Kou, Yangyu Lv, Yifei Li, Yijing Liu, Yiming Wang, Yingya Zhang, Yitong Huang, Yong Li, You Wu, Yu~Liu, Yulin Pan, Yun Zheng, Yuntao Hong, Yupeng Shi, Yutong Feng, Zeyinzi Jiang, Zhen Han, Zhi-Fan Wu, and Ziyu Liu.
\newblock Wan: Open and advanced large-scale video generative models.
\newblock \emph{arXiv preprint arXiv:2503.20314}, 2025.

\bibitem[Wang et~al.(2026{\natexlab{a}})Wang, Zheng, Chen, Li, Zhang, Xing, Zhang, Li, Qian, Yang, Dong, Hao, Ye, han, Pan, and Zhao]{latentwam}
Linbo Wang, Yupeng Zheng, Qiang Chen, Shiwei Li, Yichen Zhang, Zebin Xing, Qichao Zhang, Xiang Li, Deheng Qian, Pengxuan Yang, Yihang Dong, Ce~Hao, Xiaoqing Ye, Junyu han, Yifeng Pan, and Dongbin Zhao.
\newblock Latent-wam: Latent world action modeling for end-to-end autonomous driving, 2026{\natexlab{a}}.
\newblock URL \url{https://arxiv.org/abs/2603.24581}.

\bibitem[Wang et~al.(2026{\natexlab{b}})Wang, Yang, Bai, Zhang, Liu, Zheng, Long, Lu, and Lu]{drivejepa}
Linhan Wang, Zichong Yang, Chen Bai, Guoxiang Zhang, Xiaotong Liu, Xiaoyin Zheng, Xiao-Xiao Long, Chang-Tien Lu, and Cheng Lu.
\newblock Drive-jepa: Video jepa meets multimodal trajectory distillation for end-to-end driving, 2026{\natexlab{b}}.
\newblock URL \url{https://arxiv.org/abs/2601.22032}.

\bibitem[Wang et~al.(2024)Wang, Yu, Jiang, Lan, Shi, Chang, Kautz, Li, and Alvarez]{omnidrive}
Shihao Wang, Zhiding Yu, Xiaohui Jiang, Shiyi Lan, Min Shi, Nadine Chang, Jan Kautz, Ying Li, and Jose~M Alvarez.
\newblock Omnidrive: A holistic llm-agent framework for autonomous driving with 3d perception, reasoning and planning.
\newblock \emph{arXiv preprint arXiv:2405.01533}, 2024.

\bibitem[Wang et~al.(2026{\natexlab{c}})Wang, Xiang, Zhou, Wang, Huang, Huang, Wei, Zhou, Wang, Chen, Xu, Tan, Zhou, and Yang]{wajepa}
Xinlin Wang, Yujiao Xiang, Yuheng Zhou, Jingqi Wang, Minqing Huang, Jiajie Huang, Dongxu Wei, Tingguang Zhou, Xiyang Wang, Gong Chen, Zhi Xu, Feiyang Tan, Hangning Zhou, and Mu~Yang.
\newblock Wa-jepa: Rethinking the video jepa paradigm for world-action modeling in autonomous driving, 2026{\natexlab{c}}.
\newblock URL \url{https://arxiv.org/abs/2608.20974}.

\bibitem[Xia et~al.(2026)Xia, Li, Zhou, Yao, Xiong, Sun, Wang, Ma, Chen, Ye, et~al.]{drivelaw}
Tianze Xia, Yongkang Li, Lijun Zhou, Jingfeng Yao, Kaixin Xiong, Haiyang Sun, Bing Wang, Kun Ma, Guang Chen, Hangjun Ye, et~al.
\newblock Drivelaw: Unifying planning and video generation in a latent driving world.
\newblock In \emph{Proceedings of the IEEE/CVF Conference on Computer Vision and Pattern Recognition}, pages 39701--39712, 2026.

\bibitem[Xing et~al.(2025)Xing, Zhang, Hu, Jiang, He, Zhang, Long, and Yin]{goalflow}
Zebin Xing, Xingyu Zhang, Yang Hu, Bo~Jiang, Tong He, Qian Zhang, Xiaoxiao Long, and Wei Yin.
\newblock Goalflow: Goal-driven flow matching for multimodal trajectories generation in end-to-end autonomous driving.
\newblock In \emph{Proceedings of the Computer Vision and Pattern Recognition Conference}, pages 1602--1611, 2025.

\bibitem[Xu et~al.(2026)Xu, Zhong, Shu, Jia, Li, Bian, Zhang, Zhang, Xie, Yang, et~al.]{eponav2}
Jiawei Xu, Zhizhou Zhong, Zhijian Shu, Mingkai Jia, Mingxiao Li, Jia-Wang Bian, Qian Zhang, Kaicheng Zhang, Jin Xie, Jian Yang, et~al.
\newblock Eponav2: Driving world model with comprehensive future reasoning.
\newblock \emph{arXiv preprint arXiv:2605.14696}, 2026.

\bibitem[Xu et~al.(2024)Xu, Zhang, Xie, Zhao, Guo, Wong, Li, and Zhao]{drivegpt4}
Zhenhua Xu, Yujia Zhang, Enze Xie, Zhen Zhao, Yong Guo, Kwan-Yee~K Wong, Zhenguo Li, and Hengshuang Zhao.
\newblock Drivegpt4: Interpretable end-to-end autonomous driving via large language model.
\newblock \emph{IEEE Robotics and Automation Letters}, 2024.

\bibitem[Ye et~al.(2026)Ye, Ge, Zheng, Gao, Yu, Kurian, Indupuru, Tan, Zhu, Xiang, et~al.]{dreamzero}
Seonghyeon Ye, Yunhao Ge, Kaiyuan Zheng, Shenyuan Gao, Sihyun Yu, George Kurian, Suneel Indupuru, You~Liang Tan, Chuning Zhu, Jiannan Xiang, et~al.
\newblock World action models are zero-shot policies.
\newblock \emph{arXiv preprint arXiv:2602.15922}, 2026.

\bibitem[Zhang et~al.(2025)Zhang, Tang, Hu, Pan, Guo, Liu, Huang, Yuan, Zhang, Long, et~al.]{epona}
Kaiwen Zhang, Zhenyu Tang, Xiaotao Hu, Xingang Pan, Xiaoyang Guo, Yuan Liu, Jingwei Huang, Li~Yuan, Qian Zhang, Xiao-Xiao Long, et~al.
\newblock Epona: Autoregressive diffusion world model for autonomous driving.
\newblock In \emph{2025 IEEE/CVF International Conference on Computer Vision (ICCV)}, pages 27220--27230. IEEE, 2025.

\bibitem[Zhang et~al.(2026)Zhang, Chen, Gao, Wang, Ge, Hu, Shi, and Zhang]{onedrive}
Yiwei Zhang, Xuesong Chen, Jin Gao, Hanshi Wang, Fudong Ge, Weiming Hu, Shaoshuai Shi, and Zhipeng Zhang.
\newblock Onedrive: Unified multi-paradigm driving with vision-language-action models.
\newblock \emph{arXiv preprint arXiv:2604.17915}, 2026.

\bibitem[Zhao et~al.(2026{\natexlab{a}})Zhao, Fu, Wang, Wang, and Lu]{pwm}
Zhida Zhao, Talas Fu, Yifan Wang, Lijun Wang, and Huchuan Lu.
\newblock From forecasting to planning: Policy world model for collaborative state-action prediction.
\newblock \emph{Advances in Neural Information Processing Systems}, 38:\penalty0 134585--134611, 2026{\natexlab{a}}.

\bibitem[Zhao et~al.(2026{\natexlab{b}})Zhao, Zhou, Xu, Sun, Zhou, Li, Liang, and Bai]{simwam}
Zongchuang Zhao, Xin Zhou, Tianyang Xu, Zhengyang Sun, Kaixuan Zhou, Honglin Li, Dingkang Liang, and Xiang Bai.
\newblock Simwam: A simple world action model for end-to-end autonomous driving.
\newblock \emph{arXiv preprint arXiv:2608.07468}, 2026{\natexlab{b}}.

\bibitem[Zhou et~al.(2026)Zhou, Han, Yang, Ma, Tresp, and Knoll]{opendrivevla}
Xingcheng Zhou, Xuyuan Han, Feng Yang, Yunpu Ma, Volker Tresp, and Alois Knoll.
\newblock Opendrivevla: Towards end-to-end autonomous driving with large vision language action model.
\newblock In \emph{Proceedings of the AAAI Conference on Artificial Intelligence}, volume~40, pages 13782--13790, 2026.

\bibitem[Zhou et~al.(2025)Zhou, Cai, Zhao, Zhang, Huang, Zhou, and Ma]{autovla}
Zewei Zhou, Tianhui Cai, Seth~Z. Zhao, Yun Zhang, Zhiyu Huang, Bolei Zhou, and Jiaqi Ma.
\newblock Autovla: A vision-language-action model for end-to-end autonomous driving with adaptive reasoning and reinforcement fine-tuning, 2025.
\newblock URL \url{https://arxiv.org/abs/2506.13757}.

\bibitem[Zou et~al.(2025)Zou, Chen, Liao, Zheng, Song, Zhang, Zhang, Liu, and Wang]{diffusiondrivev2}
Jialv Zou, Shaoyu Chen, Bencheng Liao, Zhiyu Zheng, Yuehao Song, Lefei Zhang, Qian Zhang, Wenyu Liu, and Xinggang Wang.
\newblock Diffusiondrivev2: Reinforcement learning-constrained truncated diffusion modeling in end-to-end autonomous driving.
\newblock \emph{arXiv preprint arXiv:2512.07745}, 2025.

\end{thebibliography}
